%% file: acl_latex.tex
\pdfoutput=1
\documentclass[11pt]{article}

\usepackage[final]{acl}

\usepackage{times}
\usepackage{latexsym}

\usepackage[T1]{fontenc}
\usepackage[utf8]{inputenc}
\DeclareUnicodeCharacter{22C6}{\ensuremath{\star}}

\usepackage{microtype}

\usepackage{inconsolata}

\usepackage{listings}
\usepackage{natbib}
\setcitestyle{authoryear,round,semicolon}

\usepackage{graphicx}
\usepackage{hyperref}       % hyperlinks
\usepackage{url}            % simple URL typesetting
\usepackage{booktabs}       % professional-quality tables
\usepackage{amsfonts}       % blackboard math symbols
\usepackage{nicefrac}       % compact symbols for 1/2, etc.
\usepackage{xcolor}         % colors
\usepackage{todonotes}
\usepackage[linesnumbered,ruled,vlined]{algorithm2e}
\usepackage{amsmath}

\usepackage[normalem]{ulem}
\usepackage{dblfloatfix}
\usepackage{xurl}
\newcommand{\pipeline}{\texttt{seqBench}}

\title{\pipeline: A Tunable Benchmark to Quantify Sequential Reasoning Limits of LLMs}

\author{%
  M.R. Ramezanali\thanks{⋆ denotes equal contribution.} \\
  Salesforce AI\\
  Palo Alto, CA 94301 \\
  \texttt{mramezanali@salesforce.com}
  \And
  M. Vazifeh\footnotemark[1] \\
  Capital One, MIT \\
  Cambridge, MA 02143 \\
  \texttt{mvazifeh@mit.edu}
  \And
  P. Santi \\
  MIT \\
  Cambridge, MA 02143 \\
  \texttt{psanti@mit.edu} 
}

\begin{document}
\maketitle

\input{sections/abstract}
\input{sections/introduction}

\input{sections/methods}

\input{sections/results}
\input{sections/relatedwork}
\input{sections/limitations}
\input{sections/conclusions}

% \begin{ack}
% We thank the anonymous reviewers for their valuable feedback.
% \end{ack}
\clearpage
% Use the precompiled .bbl file instead of the .bib
% \bibliography{acl_latex}

\input{acl_latex.bbl}
\clearpage
\appendix
\input{sections/appendix}

\end{document}

%% file: sections/abstract.tex
\begin{abstract}
We introduce \pipeline, a parametrized benchmark for probing sequential reasoning limits in Large Language Models (LLMs) through precise, multi-dimensional control over several key complexity dimensions. \pipeline{} allows systematic variation of (1) the \emph{logical depth}, defined as the number of sequential actions required to solve the task; 
(2) the number of \emph{backtracking} steps along the optimal path, quantifying how often the agent must revisit prior states to satisfy deferred preconditions (e.g., retrieving a key after encountering a locked door); and 
(3) the \emph{noise ratio}, defined as the ratio between supporting and distracting facts about the environment. Our evaluations on state-of-the-art LLMs reveal a universal failure pattern: accuracy collapses exponentially beyond a model-specific logical depth. Unlike existing benchmarks, \pipeline{}'s fine-grained control facilitates targeted analyses of these reasoning failures, illuminating universal scaling laws and statistical limits, as detailed in this paper alongside its generation methodology and evaluation metrics. We find that even top-performing models systematically fail on \pipeline{}'s structured reasoning tasks despite minimal search complexity, underscoring key limitations in their commonsense reasoning capabilities. Designed for future evolution to keep pace with advancing models, the \pipeline{} datasets are publicly released to spur deeper scientific inquiry into LLM reasoning, aiming to establish a clearer understanding of their true potential and current boundaries for robust real-world application.
\end{abstract}

%% file: sections/introduction.tex
Large Language Models (LLMs) have shown remarkable performance \citep{NIPS2017_3f5ee243, brown2020language, lieber2021jurassic, rae2021gopher, smith2022mt530b, thoppilan2022lamda, hoffmann2022training, Du2021GLaMES, JMLR:v23:21-0998, zoph2022stmoedesigningstabletransferable} on a wide range of tasks and benchmarks spanning diverse human-like capabilities; however, these successes can obscure fundamental limitations in sequential reasoning that still persist. Arguably, reasoning captures a more pure form of intelligence, going beyond mere pattern matching or fact memorization, and is thus a critical capability to understand and enhance in AI systems. Recent studies show that state-of-the-art LLMs \citep{openAI2025o3o4mini, google2025gemini25, meta2025llama4, mistral2024large2, anthropic2025claude37} excel at complex benchmarks, yet stumble upon simple common-sense inferences trivial for an adult human \citep{nez2024alice, han2024incontextlearningelicittrustworthy, sharma2024exploring, berglund2024reversalcursellmstrained, Yang:2019:ACB}. Most existing benchmarks saturate quickly, leaving little room for fine-grained attribution studies to perform systemic probes of LLM failure modes. Consequently, a robust understanding of why and under what circumstances these models fail, especially on problems requiring sequential reasoning, remains elusive.

This gap, we argue, stems from the lack of evaluation benchmarks allowing systematic, multi-dimensional control over key independent factors that influence a task's overall reasoning difficulty. Most benchmarks \citep{GSM8K, mmlu, bigbench, babi2015, arc2018, drop2019, gpqa}, despite their evaluation merits, often do not support a systematic variation of crucial complexity dimensions. This makes it difficult to isolate the specific conditions under which reasoning in LLMs falter. For instance, discerning whether a failure is due to the length of the required reasoning chain, the necessity to revise intermediate conclusions, or the density of distracting information is often not quantitatively possible. While prompting strategies like chain-of-thought (CoT) and model scaling have boosted aggregate performance, they often obscure sharp performance cliffs that can emerge when these underlying complexity dimensions are varied independently \citep{wei2023chainofthoughtpromptingelicitsreasoning, NEURIPS2022_8bb0d291}. Without such systematic control, disentangling inherent architectural limitations from those addressable via scaling (model size, data, or compute), fine-tuning, or prompting techniques is challenging. A fine-grained understanding of these performance boundaries is crucial for developing more robust and reliable reasoning systems.

\begin{figure}[t]
  \includegraphics[width=1.0\linewidth]{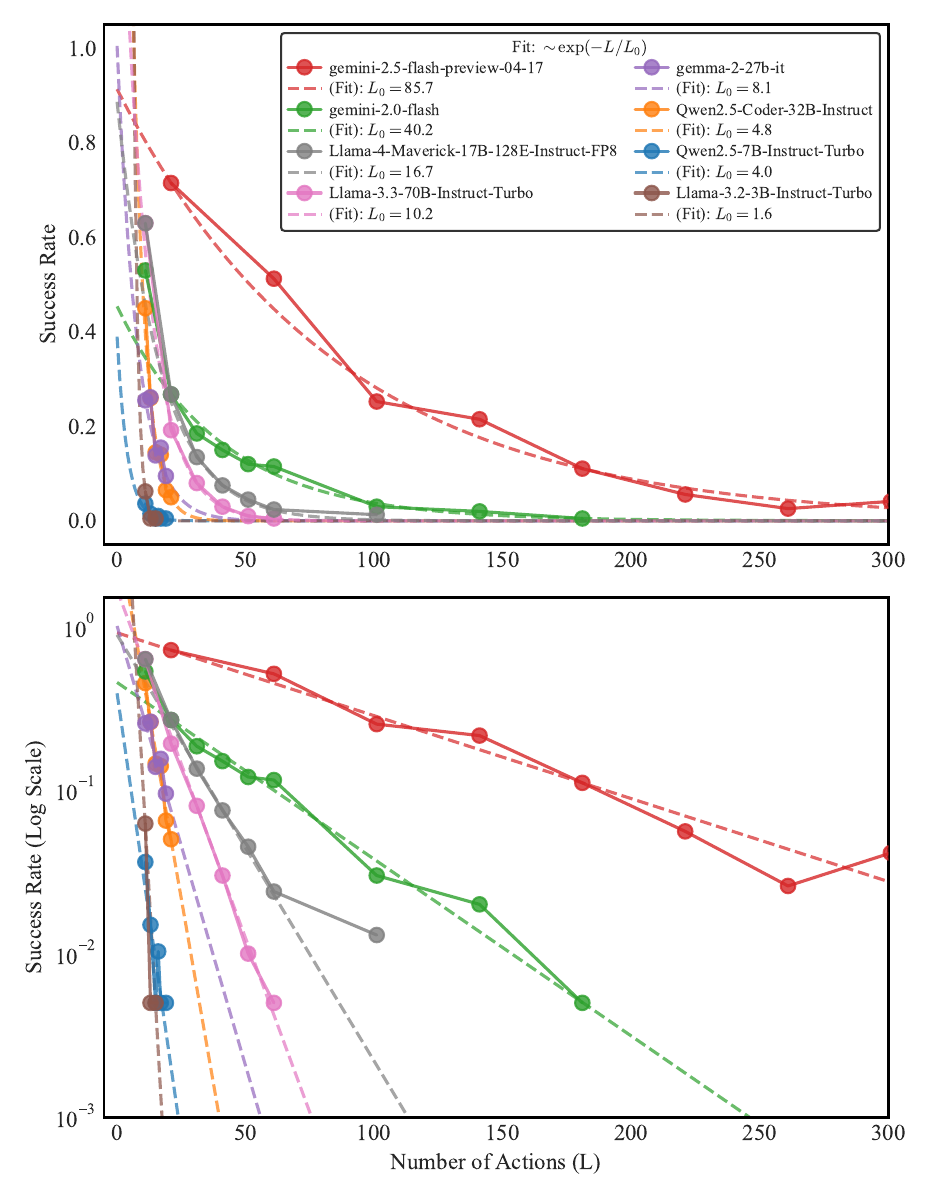}
  \centering
  \caption{Performance collapse of various models with increasing logical depth $L$ for a pathfinding task ($N,M=40, \mathcal{B}=2$ keys, Noise Ratio $\mathcal{N} = 0.0$). Success rates (Pass@1) are shown on linear (top panel) and logarithmic (bottom panel) y-axes, averaged from 5 runs/problem across 40 problems per unit $L$-bin. All evaluations used Temperature=1.0 and top-p=0.95 (Gemini-2.5-flash: 'auto' thinking). The displayed fits employ a Weighted Least Squares (WLS) \cite{wls} method on log-success rates. Weights are derived from inverse squared residuals of a preliminary Ordinary Least Squares (OLS) fit. (In the supplementary section, we have added Figure~\ref{fig:gpt5} to show a similar pattern is observed in recently released OpenAI models.)}
\label{fig:universal_vs_L}
\end{figure}

To complement recent efforts \citep{MuSR2023, tyagi2024, BabiLong2024, tang2024grasp, SPARTQ2021, PLUGH2024, mirzaee2022transfer, StepGame2022} in evaluating reasoning, and to address the need for more controlled analysis, we introduce \pipeline, a tunable benchmark designed explicitly to probe and analyze sequential reasoning capabilities in language models. The dataset comprises synthetic yet linguistically grounded pathfinding task configurations on two-dimensional grids. Solving each problem requires sequential inference over relevant and distracting structured facts. Each instance is automatically verifiable and parameterized by controllable factors that directly address the previously identified gaps: (1) logical depth (total number of actions in the ground-truth solution, reflecting the length of the reasoning chain); (2) backtracking count (number of locked-door detours on the optimal path, requiring revision of tentative solution paths); and (3) noise ratio (proportion of distracting vs. supporting facts, testing robustness to irrelevant information). Performance against these dimensions can be quantified with fine-grained metrics (e.g., via progress ratio as we define here). We observe that beyond a certain logical depth, Pass@1 success collapses to near zero for all models (see Figure~\ref{fig:universal_vs_L}). These features enable precise attribution studies of model failure modes, offering insights into the brittle boundaries of current LLM generalization.

Furthermore, the \pipeline{} benchmark is built upon a scalable data generation framework, allowing it to evolve alongside increasingly capable models to help with both model training and evaluation. Through evaluations on popular LLMs, we reveal that top-performing LLMs exhibit steep universal declines as either of the three complexity dimensions increases, while remaining comparatively robust to fact shuffle, despite the underlying logical structure being unchanged.
\paragraph{Contributions.}
Our main contributions are:
\begin{enumerate}
    \item \textbf{\pipeline{}: A Tunable Benchmark for Sequential Reasoning.} We introduce an open-source framework for generating pathfinding tasks with fine-grained, orthogonal control over logical depth, backtracking steps, and noise ratio. We also evaluate secondary factors like fact ordering (shuffle ratio; See supplementary material for details).
    \item \textbf{Comprehensive LLM Attribution Study.} Using \pipeline{}, we demonstrate the significant impact of these controlled complexities on LLM performance, revealing sharp performance cliffs in state-of-the-art models even when search complexity is minimal.
\end{enumerate}
The \pipeline{} dataset is publicly available\footnote{\url{https://huggingface.co/datasets/emnlp-submission/seqBench}} under the CC BY 4.0 license to facilitate benchmarking.

%% file: sections/methods.tex
\begin{figure}[tb!]
  \includegraphics[width=1.0\linewidth]{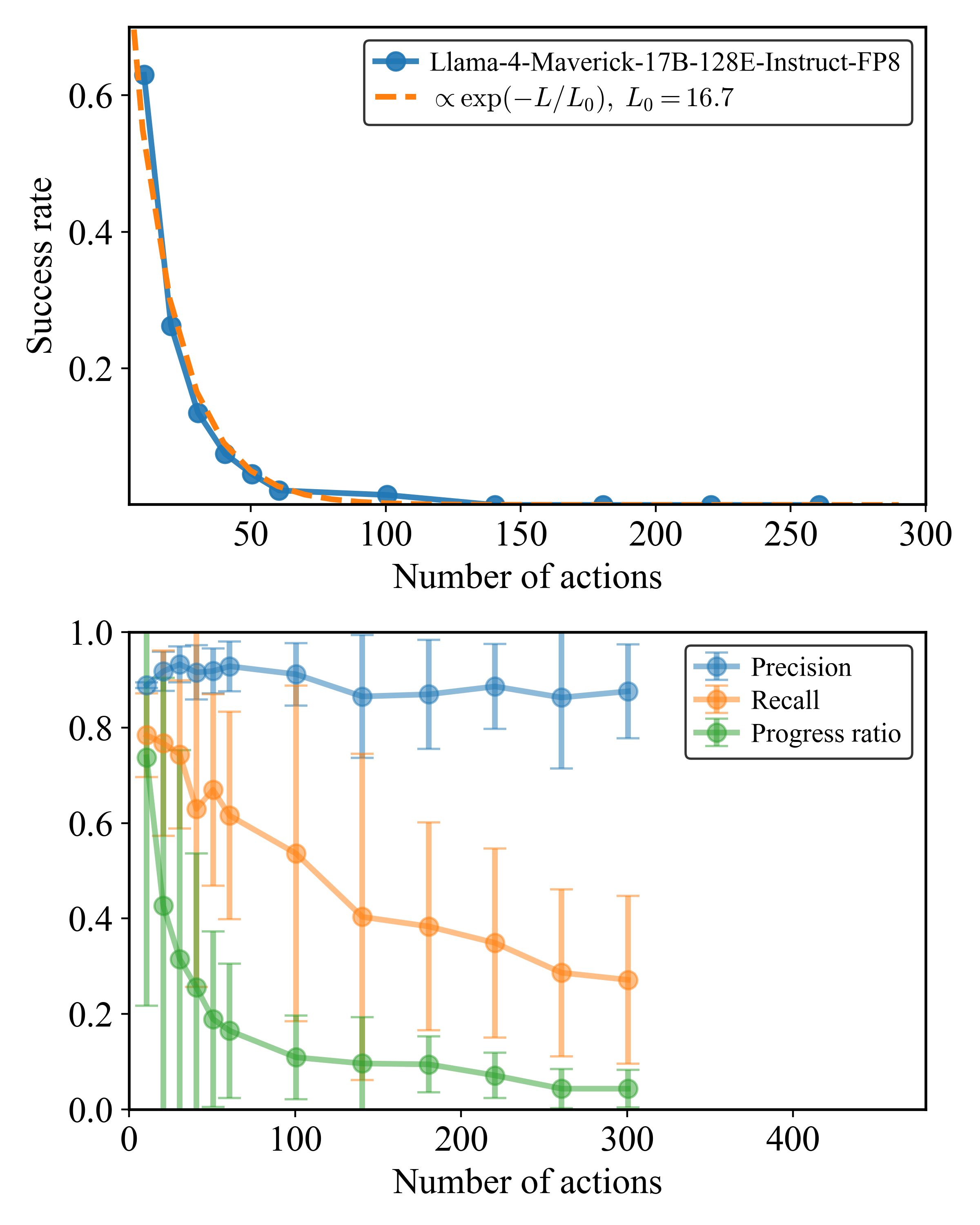}
  \centering
  \caption{On the left: Llama-4 Maverick-17B-128E-Instruct Model's performance (pass@1 success rate) versus number of actions in the ground truth path of the pathfinding problems ($N,M=40, \mathcal{B}=2$ keys, Noise Ratio $\mathcal{N} = 0.0$) is shown. This Pass@1 success rate across 5 runs per problem is averaged over the problem instances sampled from different actions count bins of width equal to 1. On the right:  The mean of progress ratio across all problems as well as mean of precision and recall is shown to highlight models gradually increasing struggle in completing the path. The Temperature is set to 1.0 and the top-p is set to 0.95 in all runs.}
  \label{fig:llama4_deepdive}
\end{figure}

\section{Methods}
\label{sec:methods}
\subsection{Dataset Generation}
The \pipeline{} dataset consists of spatial pathfinding tasks. Task instance generation, detailed below (Algorithm~\ref{alg:rewind_compact}; See Appendix~\ref{app:dataset_generation} for details), is predicated on the precise independent control of the three key complexity dimensions introduced earlier: \textbf{Logical Depth ($L$)}, \textbf{Backtracking Count ($\mathcal{B}$)}, and \textbf{Noise Ratio ($\mathcal{N}$)}. This allows the creation of instances with specific values for these parameters, enabling targeted studies of their impact on LLM reasoning.

Task instances are produced in a multi-stage process. Initially, primary generation parameters—maze dimensions ($N, M$), target backtracks ($\mathcal{B}_{\text{target}}$), and target noise ratio ($\mathcal{N}_{\text{target}}$)—are specified. An acyclic maze graph ($M_g$) is formed on an $N \times M$ grid using Kruskal’s algorithm~\citep{Kruskal}. Our "Rewind Construction" method (Algorithm~\ref{alg:rewind_compact}) then embeds $\mathcal{B}_{\text{target}}$ backtracking maneuvers by working backward from a goal to strategically place keys and locked doors, yielding the instance's actual backtracking count $\mathcal{B}$. Finally, a natural language fact list ($\mathcal{F}$) is derived from the maze, and distracting facts are added according to $\mathcal{N}_{\text{target}}$ to achieve the final noise ratio $\mathcal{N}$. The \emph{logical depth} $L$ (optimal path length) emerges from these generative steps, influenced by $N, M, \mathcal{B}_{\text{target}}$, and construction stochasticity. While $L$ is not a direct input to the generation algorithm, the process is designed to yield a wide spectrum of logical depths. Each generated instance is then precisely annotated with its emergent $L$ value, alongside its effective $\mathcal{B}$ and $\mathcal{N}$ values. This annotation effectively makes $L$ a key, selectable parameter for users of the \pipeline{} dataset, enabling them to choose or filter tasks by their desired logical depth. Our rewind construction method guarantees task solvability. The full \pipeline{} benchmark is constructed by systematically applying this instance generation process (detailed in Algorithm~\ref{alg:rewind_compact}) across a wide range of initial parameters. This includes varied grid sizes (e.g., $N \in\{5..50\}, M \approx N$) and target backtracks ($\mathcal{B}_{\text{target}}\in\{0..7\}$), yielding a large and diverse data pool. For each $(N,M,\mathcal{B}_{\text{target}})$ configuration, multiple unique base mazes are generated, to which different noise ratios (e.g., $\mathcal{N}_{\text{target}}\in\{0..1\}$) are subsequently applied. It is important to note that the algorithm constrains backtracking complexity to a simple dependency chain. In this setting, retrieving the key for each locked door involves at most one backtracking step to pick up its corresponding key, without requiring the unlocking of additional doors along the optimal path. Combined with the uniform random placement of keys, this design ensures a well-balanced distribution of backtracking difficulty across the generated instances for each logical depth $L$. Nevertheless, the same backward-in-time construction can be extended to generate tasks with higher backtracking complexity—for example, doors that require multiple keys, or intermediate doors that must be unlocked en route to other keys. Such extensions would introduce richer tree-structured dependency graphs and allow seqBench to probe model performance under more complex long-horizon reasoning regimes. The creation of this comprehensive data pool was computationally efficient, requiring approximately an hour of computation on a standard laptop while using minimal memory. The publicly released benchmark comprises a substantial collection of these generated instances, each annotated with its specific emergent logical depth $L$, effective backtracking count $\mathcal{B}$, and noise ratio $\mathcal{N}$. This rich annotation is key, enabling researchers to readily select or filter task subsets by these dimensions for targeted studies (e.g., as done for Figure~\ref{fig:universal_vs_L}, where instances were sampled into $L$-bins with other parameters fixed). For the experiments presented in this paper, specific subsets were drawn from this benchmark pool, often involving further filtering or parameter adjustments tailored to the objectives of each study; precise details for each experiment are provided in the relevant sections and figure captions. Full details on path derivation, fact compilation, and overall dataset generation parameters are provided in the Appendix~\ref{app:dataset_generation}.

\begin{algorithm}[t]
\caption{Rewind Construction of Path Skeleton}
\label{alg:rewind_compact}
\SetAlgoLined
\SetKwInOut{Input}{Input}\SetKwInOut{Output}{Output}
\Input{Grid $N \times M$, Target backtracks $\mathcal{B}$}
\Output{Maze graph $M_g$, Locked doors $\mathcal{D}_L$, Key info $\mathcal{K}_I$, Path skeleton $\Pi_S$}

$M_g \gets$ Acyclic graph on grid (Kruskal's)\;
$x \gets C_{goal} \gets$ Random goal cell in $M_g$\;
$\mathcal{D}_L, \mathcal{K}_I \gets \emptyset, \emptyset$; $b \gets 0$\;
$\Pi_S \gets [(C_{goal}, \text{GOAL})]$\; % Conceptual backward path skeleton

\While{$b < \mathcal{B}$}{
    $c_{key} \gets$ Random cell in $M_g$ accessible from $x$ (path avoids $\mathcal{D}_L$ for this step)\;
    $\pi_{seg} \gets$ Unique path in $M_g$ from $x$ to $c_{key}$\;
    \If{$\exists e \in \pi_{seg}$ such that $e \notin \mathcal{D}_L$}{
        $d \gets$ Randomly select such an edge $e$\;
        $\mathcal{D}_L \gets \mathcal{D}_L \cup \{d\}$\;
        $K_{id} \gets$ New unique key ID\;
        $\mathcal{K}_I[K_{id}] \gets \{\text{opens}: d, \text{loc}: c_{key}\}$\;
        % Prepend conceptual steps for this backtrack segment (in reverse order for $\Pi_S$)
        $\Pi_S$.prepend($(c_{key}, \text{PICKUP } K_{id})$, $(d, \text{UNLOCK } K_{id})$, $(\pi_{seg}, \text{MOVE})$)\;
        $x \gets c_{key}$; $b \gets b + 1$\;
    } \Else{ Break } % Stop if no valid placement for next backtrack
}
$\Pi_S$.prepend($(x, \text{START}))$\;
\Return $M_g, \mathcal{D}_L, \mathcal{K}_I, \Pi_S$\;
\end{algorithm}

\subsection{Prompt Construction and Model Configuration}
\label{sec:prompt_config}

Our evaluation uses a standardized prompt template with four components: (i) task instructions and action schema, (ii) three few-shot examples of increasing complexity (simple navigation, single-key, and multi-key backtracking), (iii) optional reasoning guidance, and (iv) the problem's natural-language facts. All models are queried using temperature $T{=}1.0$, nucleus sampling $p{=}0.95$, and maximum allowed setting in terms of output token limits on a per model basis. For each instance, we compute 5 independent runs to establish robust performance statistics. The complete prompt structure, shown in Figure~\ref{fig:S_prompt_template}, is provided in the Appendix~\ref{app:prompt_details}.

\subsection{Evaluation Metrics}\label{eval_method}
To analyze not just success but also \emph{how} models fail, we employ several complementary metrics. \textbf{Success Rate (Pass@1)} measures the proportion of runs where the predicted action sequence exactly matches the ground truth. The \textbf{Progress Ratio} \citep{tyagi2024}, calculated as $k/n$ (where $n$ is the total ground-truth actions and $k$ is the number correctly executed before the first error), pinpoints the breakdown position in reasoning. We also use \textbf{Precision} and \textbf{Recall}. Precision is the proportion of predicted actions that are correct, while Recall is the proportion of ground-truth actions that were correctly predicted.
Low precision indicates hallucinated actions, while low recall signifies missed necessary actions. Additionally, we visualize error locations via a \textbf{Violation Map}. This multi-faceted approach reveals each model's effective "reasoning horizon"—the maximum sequence length it can reliably traverse. Further details on all metrics and visualizations are provided in the supplementary material.

%% file: sections/results.tex
\section{Benchmarking Results}
\label{sec:results}

\begin{figure*}[t!]
  \includegraphics[width=1.0\linewidth]{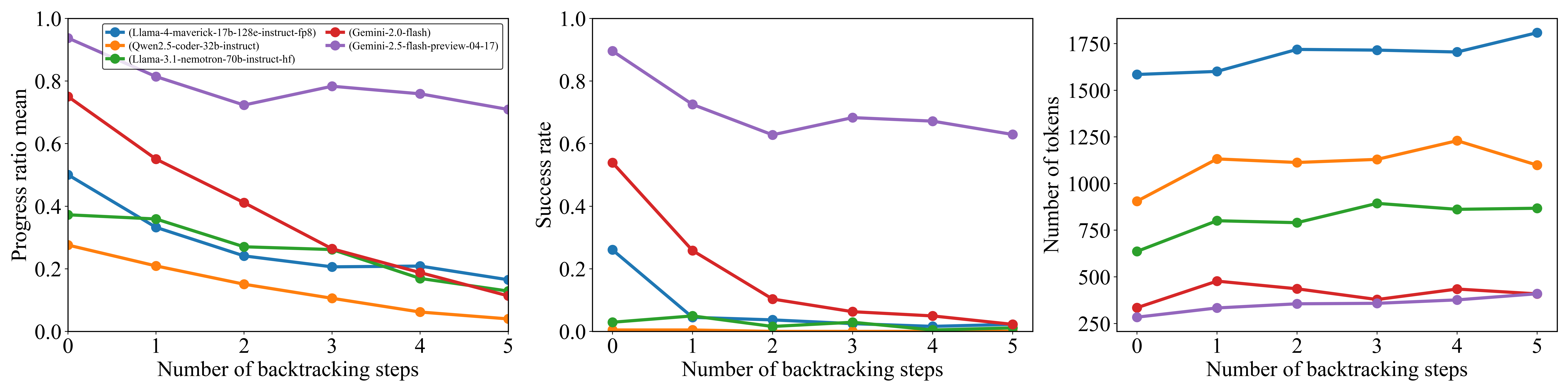}
  \centering
  \caption{Performance as a function of the number of required backtracking steps, operationalized via the number of locked doors with distributed keys along the optimal path. Holding all other complexity factors constant, all models exhibit a clear decline in both progress ratio and success rate as backtracking demands increase. Additionally, we report the corresponding rise in output token counts per model, highlighting the increased reasoning burden associated with longer dependency chains. Fixed experimental parameters in this figure are the same as those in Figure 1. (for each point 100 problems sampled from $L = [40,60]$)}
  \label{fig:vs_backtracking}
\end{figure*}

\subsection{Evaluated Models}

We evaluate a diverse set of transformer-based LLMs across different model families and parameter scales. Our analysis includes Gemini models (2.5-flash-preview, 2.0-flash), Meta's Llama family (4-Maverick-17B, 3.3-70B, 3.2-3B), Google's Gemma-2-27b, and Alibaba's Qwen models (2.5-Coder-32B, 2.5-7B). [Note: GPT-5 was released during the preparation of this paper's final version. Our analysis shows that this model exhibits the same performance degradation, as shown in Figure~\ref{fig:gpt5}]. Access to some open-weight models and benchmarking infrastructure was facilitated by platforms such as Together AI\footnote{\url{https://www.together.ai/}} and Google AI Studio\footnote{\url{https://aistudio.google.com/}}.
Problem instances for varying logical depths ($L$) were generated by sampling 40 problems for each $L$, using a fixed maze size of $40 \times 40$ and 2 keys, unless otherwise specified for specific experiments (e.g., when varying the number of keys for backtracking analysis).
All models were evaluated using the standardized prompt template (see Figure~\ref{fig:S_prompt_template}), the inference settings detailed in Section \ref{sec:prompt_config}, and a common response parsing methodology. For each task instance, we perform 5 independent runs to establish robust performance statistics, primarily analyzing Pass@1 success rates.

\subsection{Universal Performance Collapse with Increasing Logical Depth}
\label{sec:universal_collapse}

A central finding of our study is the universal collapse in reasoning performance observed across all evaluated LLMs when confronted with tasks requiring increasing sequential inference steps. As illustrated in Figure~\ref{fig:universal_vs_L}, Pass@1 success rates exhibit a consistent and sharp exponential decay as the ground-truth path length ($L$) increases. Performance rapidly approaches near-zero past a model-specific point in this decay.
To quantify and compare this exponential decay, we fit an exponential decay curve $P(L) = \exp(-L/L_0)$ to the success rates, deriving a characteristic path length $L_0$. This $L_0$ value, representing the path length at which performance drops by a factor of $e^{-1}$, serves as a robust metric for each model's sequential reasoning horizon.
Plotting success rates on a semi-logarithmic (log-y) scale against $L$ reveals an approximately linear decay trend across the evaluated regime. This log-linear relationship suggests that errors may accumulate with a degree of independence at each reasoning step, eventually overwhelming the model's capacity for coherent inference. The observed $L_0$ values vary significantly, from 85.7 for Gemini-2.5-Flash down to 1.6 for Llama-3.2-3B (Figure~\ref{fig:universal_vs_L}), underscoring a fundamental bottleneck in current transformer architectures for extended multi-step reasoning.

\subsection{Impact of Independently Controlled Complexity Dimensions}
\label{sec:complexity_dimensions}

Beyond the universal impact of logical depth ($L$) discussed in Section~\ref{sec:universal_collapse}, our benchmark's ability to independently vary key complexity dimensions allows for targeted analysis of their distinct impacts on LLM reasoning performance. We highlight the effects of noise, backtracking, and fact ordering, primarily focusing on Pass@1 success rates, mean progress ratios, and response token counts.

\begin{figure*}[t!]
  \includegraphics[width=1.0\linewidth]{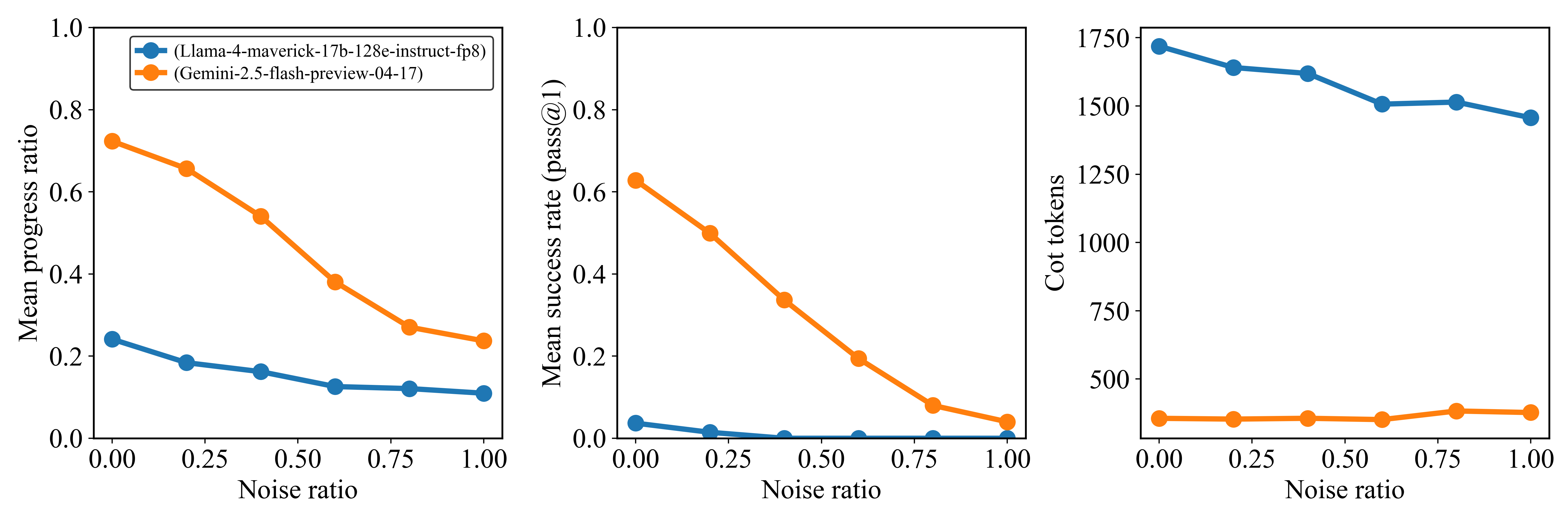}
  \centering
  \caption{Performance as a function of contextual noise for Gemini 2.5 flash and Llama-4 Maverick-17B-128E-Instruct models. As noise increases through the inclusion of distracting or irrelevant facts, both models exhibit a clear and consistent decline in performance. Fixed experimental parameters in this figure are the same as those in Figure 1 (for each point 100 problems sampled from $L = [40,60]$ and number of keys is equal to 2).}
  \label{fig:vs_noise}
\end{figure*}

\paragraph{Impact of Backtracking Requirements.}
Increasing the number of required backtracking steps—operationalized via key-door mechanisms—also leads to a clear and significant decline in Pass@1 success rates and mean progress ratios across all evaluated models as shown in Figure~\ref{fig:vs_backtracking}. Gemini 2.5 Flash-preview maintains the highest performance but still exhibits a notable drop as backtracking count increases from 0 to 5. This decline in reasoning accuracy is generally accompanied by an increase or sustained high level in the mean number of response tokens (Figure~\ref{fig:vs_backtracking}, right panel). For example, models like Llama-4 Maverick and Gemini 2.5 Flash-preview show a clear upward trend or maintain high token counts as backtracking complexity rises, reflecting the increased reasoning effort or path length articulated by the models when managing more complex sequential dependencies.

\paragraph{Sensitivity to Noise Ratio.}
Model performance is highly sensitive to the noise ratio—the proportion of distracting versus supporting facts. As demonstrated in Figure~\ref{fig:vs_noise} for Gemini 2.5 Flash and Llama-4 Maverick, increasing the proportion of irrelevant facts consistently and significantly degrades both Pass@1 success rates and mean progress ratios. For instance, Gemini 2.5 Flash's Pass@1 success rate drops from over 0.7 at zero noise to approximately 0.2 at a noise ratio of 1.0. Llama-4 Maverick, starting with lower performance, also shows a consistent decline. Interestingly, for these two models, the number of CoT (output) tokens remains relatively stable despite the increasing noise and degrading performance (Figure~\ref{fig:vs_noise}, right panel), suggesting that models do not necessarily "work harder" (in terms of output length) when faced with more distractors, but their accuracy suffers.

\paragraph{Fact Ordering (Shuffle Ratio).}
In contrast to the strong effects of noise and backtracking, shuffle ratio (entropy of fact presentation order) within the prompt appears to play a secondary role when varied in isolation. Our experiments, exemplified by the performance of Gemini 2.5 Flash and Llama-4 Maverick (see Appendix~\ref{app:dataset_figures} Figure~\ref{fig:S_vs_shuffle} for details), show that complete shuffling of facts (randomizing their presentation order without adding or removing any information) has a minimal impact on Pass@1 success rates and mean progress ratios. Output token counts also remain stable. This suggests a relative robustness to presentation order as long as all necessary information is present and distinguishable. However, as details provided in supplementary material, when high noise and high shuffle co-occur, the combined effect can be more detrimental than either factor alone, though noise remains the dominant degrading factor.

\subsection{Characterizing Key Failure Modes and Error Patterns}
\label{sec:error_pattern}

\paragraph{A Key Failure Mode: Omission of Critical Steps.}
Beyond simply taking illegal shortcuts, detailed analysis reveals that LLMs often fail by omitting critical sub-goals necessary for task completion. Figure~\ref{fig:llama4_deepdive} (bottom panel) provides a quantitative view for Llama-4 Maverick \citep{meta2025llama4}, showing that while precision generally remains high (models infrequently hallucinate non-existent rooms or facts), recall and progress ratio plummet with increasing path length ($L$). This indicates that models predominantly fail by missing necessary actions or entire crucial sub-sequences. 
For a qualitative example, even capable models like Gemini-2.5-Flash can neglect essential detours, such as collecting a required key, thereby violating sequential dependencies and rendering the task unsolvable (illustrative examples are provided in the Appendix~\ref{app:quantitative_error_analysis}; see Figures~\ref{fig:S_goodexample4040} and~\ref{fig:mistakev2}). This pattern highlights a fundamental breakdown in robust multi-step planning and execution.

\paragraph{Path-Length Dependent First Errors: The Burden of Anticipated Complexity.}
The propensity for models to make critical errors is not uniformly distributed across the reasoning process, nor is it solely a feature of late-stage reasoning fatigue. Examining the distribution of steps at which the first constraint violations occur reveals a counterintuitive pattern: as the total required path length ($L$) of a problem increases, models tend to fail more frequently \emph{even at the earliest steps} of the reasoning chain. This leftward shift in the first-error distribution also observed under increasing noise, (Appendix~\ref{app:quantitative_error_analysis}; Figures~\ref{fig:S_fail_dist_vs_L} and ~\ref{fig:S_dist_vs_noise}) contradicts a simple cumulative error model where each step carries a fixed, independent failure probability. Instead, an error at an early step (e.g., step 5) becomes substantially more likely when the model is attempting to solve an 80-step problem versus a 20-step problem. This suggests that the overall anticipated complexity of the full problem influences reasoning quality from the very outset, indicating a struggle with global planning or maintaining coherence over longer horizons, rather than just an accumulation of local errors. This phenomenon may help explain why prompting techniques that decompose long problems into smaller, manageable sub-problems often succeed.

\subsection{Disparity: Information Retention vs. Reasoning Capacity}
\label{sec:disparity}

On \pipeline{} tasks, this disparity is quantitatively striking. While modern LLMs boast million-token contexts, their effective sequential reasoning depth typically remains on the order of hundreds of actions (Figure~\ref{fig:universal_vs_L}). This functional limit, even at several hundred actions (e.g., 300 actions, with each like \texttt{('move\_to', 'A12')} being ~5-7 tokens, totaling ~1.5k-2.1k tokens), still consumes a minute fraction of their nominal context. Consequently, the ratio of context capacity to reasoning tokens often spans from several hundred-fold (e.g., ~500:1 for ~300 actions consuming ~2k tokens within a 1M context) to potentially higher values given fewer limiting actions or larger model contexts. This striking gap suggests that while transformers can store and retrieve vast information, their ability to reliably chain it for coherent, multi-step inference appears surprisingly constrained.

\subsection{Challenging the Conventional Performance Hierarchy}
\label{sec:hierarchy}

While metrics like average $L_0$ provide a general ranking of model capabilities, our fine-grained analysis reveals instances that challenge a simple linear performance hierarchy. Scatter plots of progress ratios across different models on identical tasks (see Appendix~\ref{app:dataset_figures} Figure~\ref{fig:S_progress_vs_progress}) show intriguing cases where models with lower overall $L_0$ values (i.e., typically weaker models) occasionally solve specific complex problems perfectly, while models with higher average $L_0$ values fail on those same instances. These performance inversions suggest that sequential reasoning failures may not solely stem from insufficient scale (parameters or general training) but could also arise from more nuanced reasoning limitations.

%% file: sections/relatedwork.tex
\section{Related Work}

Recent advancements in benchmarks evaluating sequential reasoning capabilities of LLMs have illuminated various strengths and limitations across different dimensions of complexity. These benchmarks typically differ in how they isolate and quantify reasoning challenges, such as logical deduction, retrieval difficulty, combinatorial complexity, and sensitivity to irrelevant information. ZebraLogic \citep{ZebraLogic2025}, for instance, targets formal deductive inference through logic-grid puzzles framed as constraint-satisfaction problems \citep{csp2008}. While valuable for probing deduction, its core methodology leads to a search space that grows factorially with puzzle size \citep{Sempolinski2009AutomaticSO}. This makes it challenging to disentangle intrinsic reasoning failures from the sheer combinatorial complexity of the search. As the ZebraLogic authors themselves acknowledge: ``\textit{solving ZebraLogic puzzles for large instances may become intractable... the required number of reasoning tokens may increase exponentially with the size of the puzzle.}'' This inherent characteristic means that for larger puzzles, performance is primarily dictated by the manageability of the search space rather than the limits of sequential reasoning depth. GridPuzzle \citep{tyagi2024} complements this by providing a detailed error taxonomy for grid puzzles, focusing on \textit{what} kinds of reasoning mistakes LLMs make. However, like ZebraLogic, it doesn't offer independent control over key complexity dimensions such as logical depth, backtracking needs, or noise, separate from the puzzle's inherent search complexity.

Other benchmarks conflate reasoning with different cognitive demands. BABILong \citep{BabiLong2024} tests models on extremely long contexts (up to 50M tokens), primarily assessing the ability to retrieve "needles" (facts) from a "haystack" (distracting text that does not contribute to solving the task). While valuable for evaluating long-context processing, this design makes it hard to disentangle retrieval failures from reasoning breakdowns, as performance is often dictated by finding the relevant information rather than reasoning over it. MuSR \citep{MuSR2023} embeds reasoning tasks within lengthy narratives (e.g., murder mysteries), mixing information extraction challenges with complex, domain-specific reasoning structures. This realism obscures which specific aspect—extraction or reasoning depth—causes model failures. Dyna-bAbI \citep{Tamari2021} offers a dynamic framework for compositional generalization but focuses on qualitative combinations rather than systematically varying quantitative complexity metrics needed to find precise failure points.

Spatial reasoning benchmarks, while relevant, also target different aspects. GRASP \citep{tang2024grasp} assesses practical spatial planning efficiency (like obstacle avoidance) in 2D grids, a different skill than the abstract sequential reasoning \pipeline{} isolates. SPARTQA \citep{SPARTQ2021} focuses on specialized spatial relational complexity (transitivity, symmetry) using coupled dimensions, preventing independent analysis of factors like path length. SpaRTUN \citep{mirzaee2022transfer} uses synthetic data primarily for transfer learning in Spatial Question Answering (SQA), aiming to improve model performance rather than serve as a diagnostic tool with controllable complexity. Similarly, StepGame \citep{StepGame2022} demonstrates performance decay with more reasoning steps in SQA but lacks the fine-grained, orthogonal controls over distinct complexity factors provided by \pipeline{}.

In contrast, \pipeline{} takes a targeted diagnostic approach. By deliberately simplifying the spatial environment to minimize search complexity, it isolates sequential reasoning. Its core contribution lies in the independent, fine-grained control over (1) logical depth (the number of sequential actions required to solve the task), (2) backtracking count (the number of backtracking steps along the optimal path), and (3) noise ratio (the ratio of supporting to distracting facts). This orthogonal parameterization allows us to precisely pinpoint \textit{when} and \textit{why} sequential reasoning capabilities degrade, revealing fundamental performance cliffs even when search and retrieval demands are trivial. \pipeline{} thus offers a complementary tool for understanding the specific limitations of sequential inference in LLMs.

%% file: sections/limitations.tex
\section{Limitations}
\label{sec:limitations_future_work}

While \pipeline{} offers precise control over key reasoning complexities, our study has limitations that open avenues for future research:

\begin{enumerate}
    \item \textbf{Generalizability and Task Design Fidelity:}
    Our current findings are rooted in synthetic spatial pathfinding tasks. While this allows for controlled experimentation, future work must extend \pipeline{}'s methodology to more diverse reasoning domains (e.g., mathematical proofs) and incorporate greater linguistic diversity (e.g., ambiguity) to assess the broader applicability of the observed phenomena of performance collapse (quantified by $L_0$) and failure patterns. Moreover, this work did not investigate whether similar failure modes arise when the problem is also presented visually (e.g., as maze images). Multimodal capabilities could influence spatial reasoning outcomes, and we have already extended the benchmark by releasing maze image generation code alongside the HuggingFace dataset. This dataset can also be used to help train multimodal reasoning models.

    \item \textbf{Model Scope and Understanding Deeper Failure Dynamics}:
    Our current evaluation, while covering diverse public models, should be expanded to a wider array of LLMs—including recent proprietary and newer open-source variants (e.g., GPT, Claude, DeepSeek series)—to rigorously assess the universality of our findings on the characteristic length $L_0$ and failure patterns. Furthermore, while \pipeline{} effectively characterizes how reasoning performance degrades with logical depth (i.e., by determining $L_0$), two complementary research thrusts are crucial for understanding \emph{why}. First, systematic investigation is needed to disentangle how $L_0$ is influenced by factors such as model architecture, scale (parameters, training data, compute), fine-tuning strategies, and inference-time computation (e.g., chain-of-thought depth). Second, deeper analysis is required to explain the precise mechanisms underlying the observed exponential performance collapse characterized by $L_0$ and to account for other non-trivial error patterns, such as path-length dependent first errors. Additionally, the evaluation presented here does not consider how agentic systems capable of tool use perform as the reasoning complexity is tuned across various dimensions. Exploring such setups, where the LLM can externalize sub-problems, invoke tools, or backtrack programmatically, could provide valuable insights into whether the same exponential failure modes persist. In particular, one can define sequential problems where the degree of backtracking or sequential tool use can be systematically varied, and to test whether similar performance drop emerge as the dependency chain grows. We highlight this as a promising direction for future research.

\item \textbf{Impact of Prompting:}
    Our current study employed standardized prompts and inference settings. A crucial next step is a robust sensitivity analysis to determine overall decay behavior are influenced by different prompting strategies (e.g., zero-shot vs. few-shot, decomposition techniques), varied decoding parameters (temperature, top-p), and interactive mechanisms such as self-verification or self-correction. Investigating the potential of these techniques to mitigate the observed sequential inference failures, particularly given \pipeline{}'s minimal search complexity, remains a key avenue for future research.

\end{enumerate}
Addressing these points by leveraging frameworks like \pipeline{} will be vital for developing LLMs with more robust and generalizable sequential reasoning capabilities, and for understanding their fundamental performance limits.

%% file: sections/conclusions.tex
\section{Conclusion}
\label{sec:conclusion}

We introduced \pipeline, a novel benchmark framework designed for the precise attribution of sequential reasoning failures in Large Language Models. \pipeline{}'s core strength lies in its unique capability for fine-grained, independent control over fundamental complexity dimensions; most notably, logical depth ($L$), backtracking requirements, and noise ratio, its provision of automatically verifiable solutions, and critically minimizing confounding factors like search complexity. This design allows \pipeline{} to isolate and rigorously evaluate the sequential inference capabilities of LLMs, enabling the automatic quantification of fine-grained performance metrics (such as progress ratio) and providing a clear lens into mechanisms often obscured in most other benchmarks. The framework's inherent scalability and open-source nature position it as a durable tool for assessing and driving progress in current and future generations of models, ultimately aiming to enhance their utility for complex, real-world problems that often span multiple domains.
Our comprehensive evaluations using \pipeline{} reveal that reasoning accuracy consistently collapses exponentially with increasing logical depth across a diverse range of state-of-the-art LLMs. This collapse is characterized by a model-specific parameter $L_0$ (Section~\ref{sec:universal_collapse}), indicating an inherent architectural bottleneck in maintaining coherent multi-step inference. In alignment with the goal of advancing NLP's reach and fostering its responsible application in other fields by offering this precise analysis, \pipeline{} provides a valuable resource. It encourages a shift beyond aggregate benchmark scores towards a more nuanced understanding of model capabilities, an essential step for rigorously assessing the true impact and potential risks of applying LLMs in new domains. The insights gleaned from \pipeline{} can inform both NLP developers in building more robust models, and experts in other disciplines in setting realistic expectations and co-designing NLP solutions that are genuinely fit for purpose. Targeted improvements, guided by such fundamental understanding, are key to enhancing the robustness of sequential reasoning, making LLMs more reliable partners in interdisciplinary endeavors.
Future work should leverage these insights to develop models that can overcome the observed performance cliffs and extend their effective reasoning horizons, thereby unlocking their transformative potential in diverse interdisciplinary applications—such as navigating complex scientific literature, supporting intricate legal analysis, or enabling robust multi-step planning in critical autonomous systems. Focusing on commonsense reasoning is paramount for NLP to achieve transformative societal impact, moving beyond incremental improvements to genuine breakthroughs.

%% file: sections/appendix.tex
\section*{Appendices}
\appendix
%----------------------------------------------------------------------
\section{Dataset Generation Details}
\label{app:dataset_generation}

The \pipeline{} benchmark generates pathfinding tasks by systematically controlling several complexity dimensions. As described in Section~\ref{sec:methods} (main paper), Algorithm~\ref{alg:rewind_compact} is central to this process. This appendix provides further details on the generation phases, natural language encoding of tasks, and specific dataset parameters.

\subsection{Generation Phases}
\begin{figure}[hbt!]
  \centering
  \includegraphics[width=1.0\linewidth]{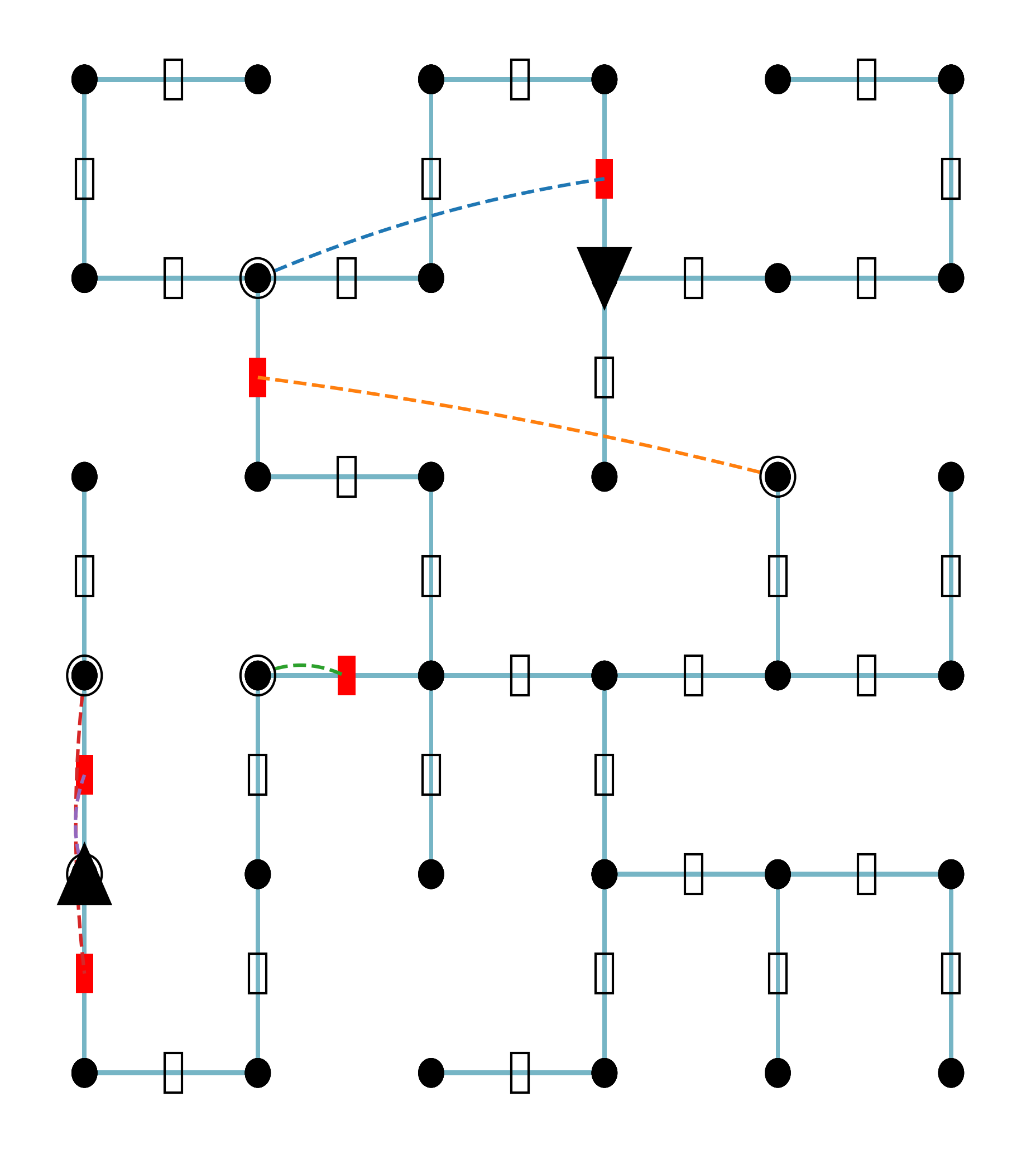}
  \caption{Example visualization of a $6 \times 6$ \pipeline{} maze instance. Red rectangles denote locked doors, dashed lines indicate the locations of keys corresponding to those doors, and triangles mark the start (upward-pointing) and goal (downward-pointing) positions. This illustrates the spatial nature of the tasks.}
  \label{fig:S_compath_viz}
\end{figure}

The generation process, guided by Algorithm~\ref{alg:rewind_compact}, involves three main phases:

\begin{enumerate}
  \item \textbf{Base Maze Construction:} An initial $N \times M$ grid is populated, and an acyclic maze graph ($M_g$) is formed using Kruskal's algorithm~\citep{Kruskal}. This ensures a simply connected environment where a unique path exists between any two cells if all internal "walls" (potential door locations) were open. The overall process results in maze instances like the one visualized in Figure~\ref{fig:S_compath_viz}.

  \item \textbf{Rewind Construction for Path Skeleton and Key/Door Placement:}
  This phase implements the "Rewind Construction" (Algorithm~\ref{alg:rewind_compact} in the main paper). Starting from a randomly selected goal cell ($C_{goal}$), the algorithm works backward to define a solvable path skeleton ($\Pi_S$). It iteratively:
    \begin{enumerate}
        \item Selects a cell $c_{key}$ that would be a preceding point on a path towards the current cell $x$ (initially $C_{goal}$).
        \item Identifies the unique path segment $\pi_{seg}$ in $M_g$ from $x$ to $c_{key}$.
        \item Randomly selects an edge $d$ on this segment $\pi_{seg}$ to become a locked door. This edge $d$ is added to the set of locked doors $\mathcal{D}_L$.
        \item A new unique key $K_{id}$ is conceptually placed at $c_{key}$, and its information (which door it opens, its location) is stored in $\mathcal{K}_I$.
        \item The conceptual steps (moving along $\pi_{seg}$, unlocking door $d$ with $K_{id}$, picking up $K_{id}$ at $c_{key}$) are prepended (in reverse logical order) to the path skeleton $\Pi_S$.
        \item The current cell $x$ is updated to $c_{key}$, and the process repeats until the target number of backtracks ($\mathcal{B}$) is achieved or no valid placements remain.
    \end{enumerate}
  This backward construction ensures solvability and controlled backtracking complexity. The final agent starting position is the cell $x$ at the end of this phase.

  \item \textbf{Fact Compilation and Noise Injection:}
  Based on the final maze structure ($M_g, \mathcal{D}_L, \mathcal{K}_I$), a set of natural language facts $\mathcal{F}$ is compiled. This includes facts describing room connections, key locations, and door states. Distracting facts are then introduced based on the target noise ratio $\mathcal{N}$. These distractors might describe non-existent connections, spurious keys, or misleading adjacencies, chosen to be plausible yet incorrect.
\end{enumerate}

\subsection{Natural Language Encoding}
Each task instance is translated into a set of atomic natural language facts. We use a consistent templating approach:
\begin{itemize}
  \item \textbf{Room Connections:} "Room A1 and B1 are connected by an open door."
  \item \textbf{Locked Connections:} "Room C3 and D3 are connected by a closed and locked door."
  \item \textbf{Key Requirements:} "The locked door between C3 and D3 requires key 5." (Key IDs are simple integers).
  \item \textbf{Key Placements:} "Key 5 is in room E4." (Room IDs use spreadsheet-like notation, e.g., A1, B2).
  \item \textbf{Starting Position:} "Bob is in room A2."
  \item \textbf{Goal Position:} "Alice is in room D5."
\end{itemize}
The full set of facts for a given problem constitutes its description.

\subsection{Dataset Parameters and Scope}
The \pipeline{} dataset was generated using the following parameter ranges based on the generation configuration:
\begin{itemize}
\item \textbf{Grid Sizes ($N \times M$):} $N \times M$ where $N$ and $M$ range from 5 to 50 (e.g., [5,5], [6,6], ..., [50,50]), with $M = N$ for all configurations.
\item \textbf{Target Backtracking Steps ($\mathcal{B}$):} Values from 0 to 7. This controls the number of key-door mechanisms deliberately placed on the optimal path.
\item \textbf{Noise Ratio ($\mathcal{N}$):} Values from $0.0$ (no distracting facts) to $1.0$ (equal number of supporting and distracting facts), typically in increments of $0.2$.\item \textbf{Instances per Configuration:} For each primary configuration, defined by a specific grid size ($N,M$) and a specific target backtracking step count ($\mathcal{B} \in \{0..7\}$), 400 unique base maze instances were generated.
\item \textbf{Logical Depth ($L$):} As an emergent property, $L$ varies. Experiments typically select problems from these generated instances that fall into specific $L$ bins (e.g., $L \in [10,11), [11,12), \ldots$).
\end{itemize}
This generation pipeline, leveraging the described parameter ranges and variations, can produce a vast and diverse set of problem instances. The publicly released \pipeline{} dataset, used for the analyses in this paper (see main paper for access link), comprises 7,079 such curated instances. This collection offers a rich resource for studying the combined effects of the controlled complexity dimensions.

%----------------------------------------------------------------------
\section{Prompt Design and Model Configuration Details}
\label{app:prompt_details}

\begin{figure*}[tb!]
\centering
\includegraphics[width=\linewidth]{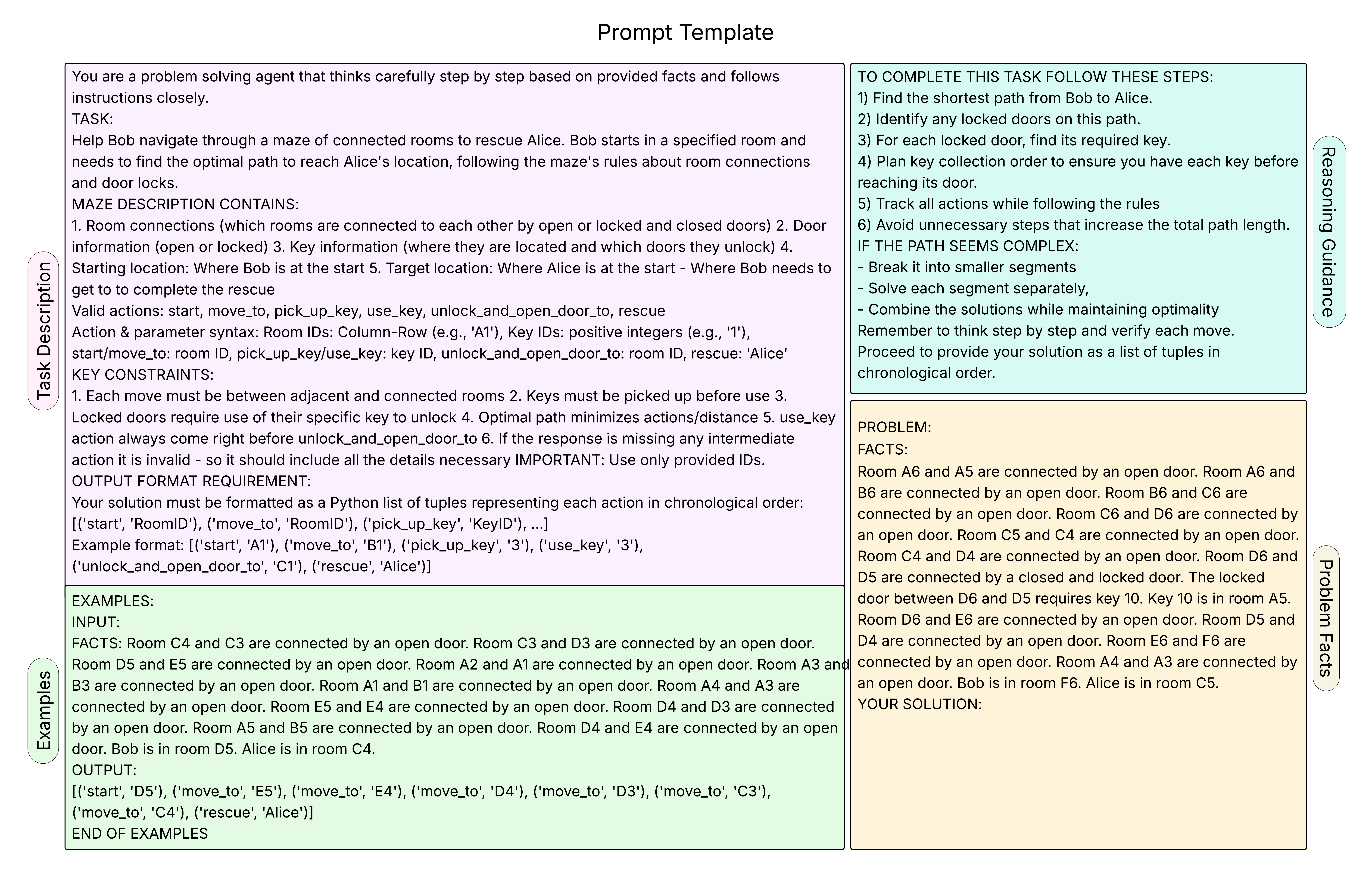}
\caption{The complete prompt structure passed to the LLMs. This includes: Component 1 (System Instructions and Task Definition), one of the three Few-Shot Examples (Component 2, specifically a simple navigation task), Component 3 (Reasoning Guidance), and an illustration of where the Problem Instance Facts (Component 4) are inserted. For clarity and completeness, the full verbatim text for all three few-shot examples (Component 2) is provided in~\ref{fig:all_few_shot_examples}.}
\label{fig:S_prompt_template}
\end{figure*}

This appendix provides the complete details of the prompt structure and model configurations used for evaluating LLMs on the \pipeline{} benchmark. The overall prompt, illustrated in Figure~\ref{fig:S_prompt_template}, concatenates four main components which are detailed below.

\subsection{Overall Prompt Components}
The prompt presented to the LLMs consists of the following components:

\begin{enumerate}
    \item \textbf{System Instructions and Task Definition (Component 1):} Outlines the agent's task, the structure of the maze description, valid actions and their syntax, key operational constraints, and the required output format.
    \item \textbf{Few-Shot Examples (Component 2):} Three examples are provided to illustrate the task, ranging in complexity. One of these examples (a simple navigation task) is detailed in Figure~\ref{fig:S_prompt_template}. The verbatim text for all three examples is provided in Figure~\ref{fig:all_few_shot_examples} for completeness.
    \item \textbf{Reasoning Guidance and Self-Assessment (Component 3):} Offers step-by-step algorithmic tips for solving the task and requests the model to provide a self-assessment of its confidence and the perceived difficulty of the instance.
    \item \textbf{Problem Instance Facts (Component 4):} The specific natural language facts describing the current maze configuration for the task instance. As illustrated in Figure~\ref{fig:S_prompt_template}, these facts are appended after the preceding components and are followed by the line "YOUR SOLUTION:" to prompt the model. These facts are generated using the templates described in Appendix~\ref{app:dataset_generation}.
\end{enumerate}

\begin{figure*}[tb!] 
\begin{enumerate}
    \item \textbf{Example 1 (Simple Navigation):}
    This example, as shown in Figure~\ref{fig:S_prompt_template}, involves navigating a maze with only open doors.
    {\small
    \begin{lstlisting}
EXAMPLE:
INPUT:
Maze Structure: Room C4 and C3 are connected by an open door. Room C3 and D3 are connected by an open door. Room D5 and E5 are connected by an open door. Room A2 and A1 are connected by an open door. Room A3 and B3 are connected by an open door. Room A1 and B1 are connected by an open door. Room A4 and A3 are connected by an open door. Room E5 and E4 are connected by an open door. Room D4 and D3 are connected by an open door. Room A5 and B5 are connected by an open door. Room D4 and E4 are connected by an open door. Bob is in room D5. Alice is in room C4.
OUTPUT:
Solution: [('start', 'D5'), ('move_to', 'E5'), ('move_to', 'E4'), ('move_to', 'D4'), ('move_to', 'D3'), ('move_to', 'C3'), ('move_to', 'C4'), ('rescue', 'Alice')]
    \end{lstlisting}
    }

    \item \textbf{Example 2 (Single-Key Backtracking):}
    This example introduces a single locked door and a corresponding key.
    {\small
    \begin{lstlisting}
EXAMPLE:
INPUT:
Maze Structure: Room A1 and A2 are connected by an open door. Room A2 and B2 are connected by an open door. Room B1 and B2 are connected by an open door. Room B1 and C1 are connected by an open door. Room C1 and C2 are connected by a closed and locked door. Door between C1 and C2 requires key 1. Key 1 is in room A2. Bob is in room A1. Alice is in room C2.
OUTPUT:
Solution: [('start', 'A1'), ('move_to', 'A2'), ('pick_up_key', '1'), ('move_to', 'B2'), ('move_to', 'B1'), ('move_to', 'C1'), ('use_key', '1'), ('unlock_and_open_door_to', 'C2'), ('move_to', 'C2'), ('rescue', 'Alice')]
    \end{lstlisting}
    }

    \item \textbf{Example 3 (Multi-Key Backtracking):}
    This example presents a more complex scenario with multiple locked doors and keys, requiring more extensive backtracking.
    {\small
    \begin{lstlisting}
EXAMPLE:
INPUT:
Maze Structure: Room B5 and B4 are connected by a closed and locked door. The locked door between B5 and B4 requires key 3. Key 3 is in room B5. Room B5 and C5 are connected by a closed and locked door. The locked door between B5 and C5 requires key 16. Key 16 is in room C5. Room B4 and C4 are connected by an open door. Room C4 and C3 are connected by an open door. Room C3 and D3 are connected by a closed and locked door. The locked door between C3 and D3 requires key 10. Key 10 is in room C4. Room D5 and D4 are connected by an open door. Room D4 and D3 are connected by an open door. Room A5 and B5 are connected by an open door. Bob is in room C5. Alice is in room D5.
OUTPUT:
Solution: [('start', 'C5'), ('pick_up_key', '16'), ('use_key', '16'), ('unlock_and_open_door_to', 'B5'), ('move_to', 'B5'), ('pick_up_key', '3'), ('use_key', '3'), ('unlock_and_open_door_to', 'B4'), ('move_to', 'B4'), ('move_to', 'C4'), ('pick_up_key', '10'), ('move_to', 'C3'), ('use_key', '10'), ('unlock_and_open_door_to', 'D3'), ('move_to', 'D3'), ('move_to', 'D4'), ('move_to', 'D5'), ('rescue', 'Alice')]
    \end{lstlisting}
    }
\end{enumerate}
\caption{Few-shot examples provided to guide the LLMs in the maze-solving task. These examples demonstrate simple navigation, single-key backtracking, and multi-key backtracking scenarios. The three examples illustrate increasing levels of complexity.}
\label{fig:all_few_shot_examples}
\end{figure*}

\subsection{Evaluation Metrics and Error Analysis Details}
This section provides further details on specific aspects of our evaluation metrics and observed error categories, complementing the overview of metrics in Section~\ref{sec:methods} of the main paper and the discussion of failure modes in Section~\ref{sec:results} of the main paper.

\paragraph{Observed Violation Categories.}
Failures in model solutions on \pipeline{} tasks can be categorized into several types. Understanding these categories is crucial for interpreting model performance and failure modes. Key types of violations observed include:
\begin{itemize}
    \item Adjacency errors (e.g., attempting to move between unconnected rooms).
    \item Locked door errors (e.g., navigating through locked doors without the correct key or without unlocking them).
    \item Key usage errors (e.g., attempting to use keys not yet collected, or using the wrong key for a door).
    \item Path inefficiency (e.g., taking unnecessary detours or redundant actions; while not always a hard violation that stops progress, this contributes to solutions not matching the optimal path and thus failing Pass@1).
    \item Missed critical actions (e.g., failing to pick up a necessary key or unlock a required door). This is a key failure mode discussed in the main paper (Section~\ref{sec:error_pattern}) and is often reflected in metrics like low recall or a low progress ratio if the omission occurs early and prevents further correct steps.
\end{itemize}
Identifying these distinct categories of errors provides a more granular understanding of why models fail on sequential reasoning tasks and helps in the interpretation of aggregate performance metrics reported in the main paper.

\subsection{Violation Map: Qualitative Examples of Model Failures}
\begin{figure*}[hbt!]
  \centering
  \includegraphics[width=1.0\linewidth]{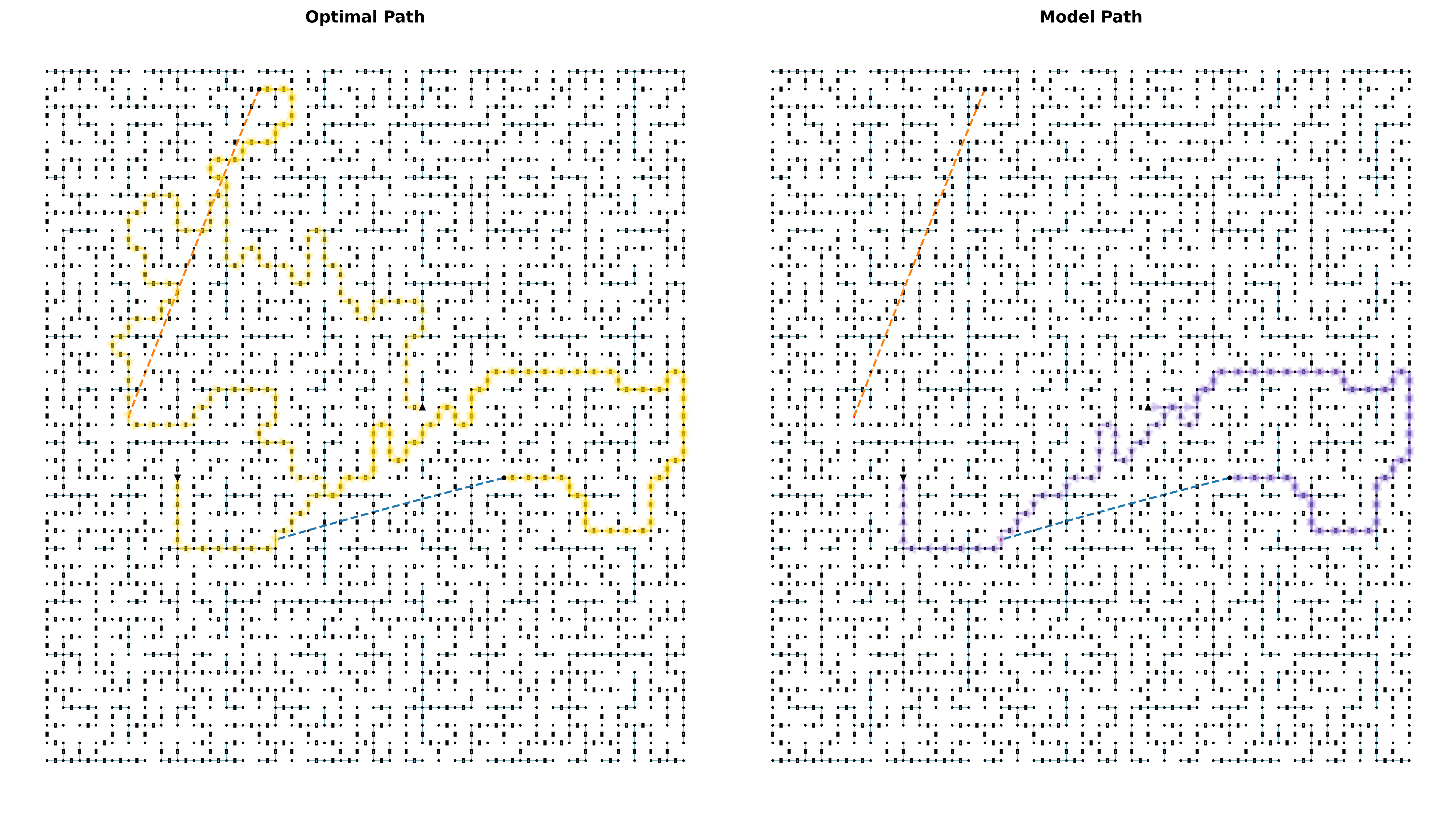}
  \caption{Illustrative failure case for Gemini-2.5-Flash on a 40x40 task with 2 locked doors on the optimal path. Left: Optimal path (yellow). Right: Model's generated path showing an illegal adjacency jump (red arrow), bypassing multiple rooms and a locked door, despite only supporting facts being provided. This highlights a breakdown in multi-step planning.}
  \label{fig:S_goodexample4040}
\end{figure*}

\begin{figure*}[hbt!]
  \includegraphics[width=1.0\linewidth]{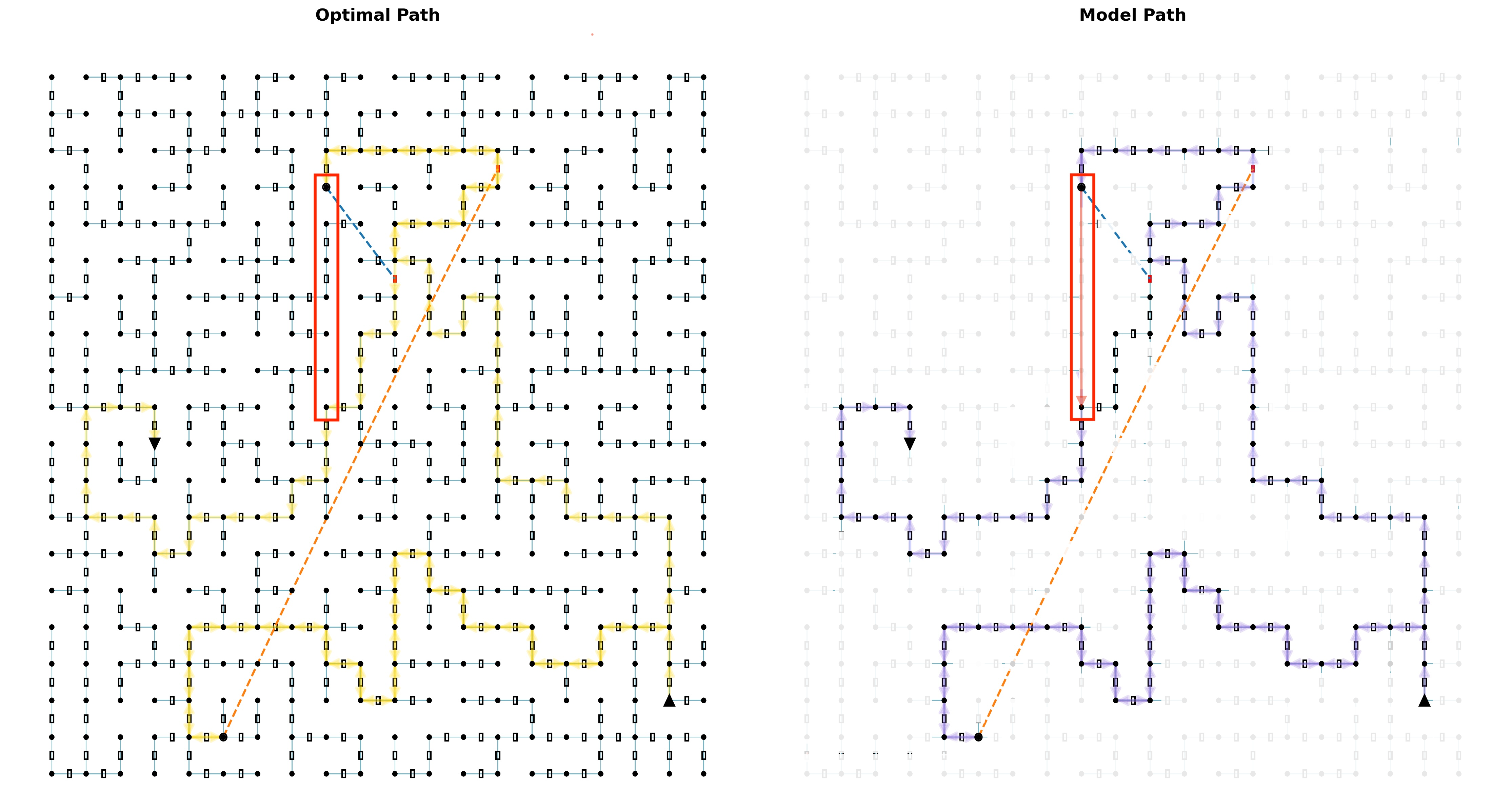}
  \centering
  \caption{Illustrative failure case of an 'adjacency error' in model-generated pathfinding on a 20x20 task with 2 locked doors on the optimal path. The left panel displays the optimal path (yellow) to the target (triangle). The right panel shows a suboptimal path (purple) generated by the model. This example highlights a common error where, after a sequence of actions (in this scenario, following a key acquisition), the model fails to navigate through valid connections. Instead, it attempts to 'jump' directly between two unconnected rooms. This violation of room adjacency constraints is a key challenge in model performance.}
  \label{fig:mistakev2}
\end{figure*}

This section provides qualitative examples of characteristic model failures to illustrate common error types. These examples visually support the discussion of failure modes in the main paper (Section~\ref{sec:error_pattern}, "A Key Failure Mode: Omission of Critical Steps"). Figure~\ref{fig:S_goodexample4040} illustrates a significant error by Gemini-2.5-Flash on a complex task, where the model generates an illegal path, bypassing necessary steps and locked doors. This exemplifies a breakdown in multi-step planning. Additionally, Figure~\ref{fig:mistakev2} shows another common 'adjacency error,' where a model attempts to jump between unconnected rooms. This type of error reveals a critical lapse in grounding its generated actions within the spatial adjacencies explicitly stated by the task's input facts.

\subsection{Quantitative Analysis of Error Patterns}
\label{app:quantitative_error_analysis}
To understand how and when models begin to fail within a reasoning sequence, we analyze the distribution of the first violation step. We record the time step at which the initial violation occurs in a model's generated path. Aggregating this step-indexed data across multiple instances allows us to create temporal distributions of errors. These distributions help determine whether errors tend to cluster early in the reasoning process (potentially indicating issues with initial planning or understanding of the overall problem complexity) or accumulate later (suggesting difficulties in maintaining long chains of inference or context). This analysis complements the discussion in the main paper (Section~\ref{sec:error_pattern}, "Path-Length Dependent First Errors: The Burden of Anticipated Complexity").

\begin{figure*}[hbt!]
  \centering
  \includegraphics[width=1.0\linewidth]{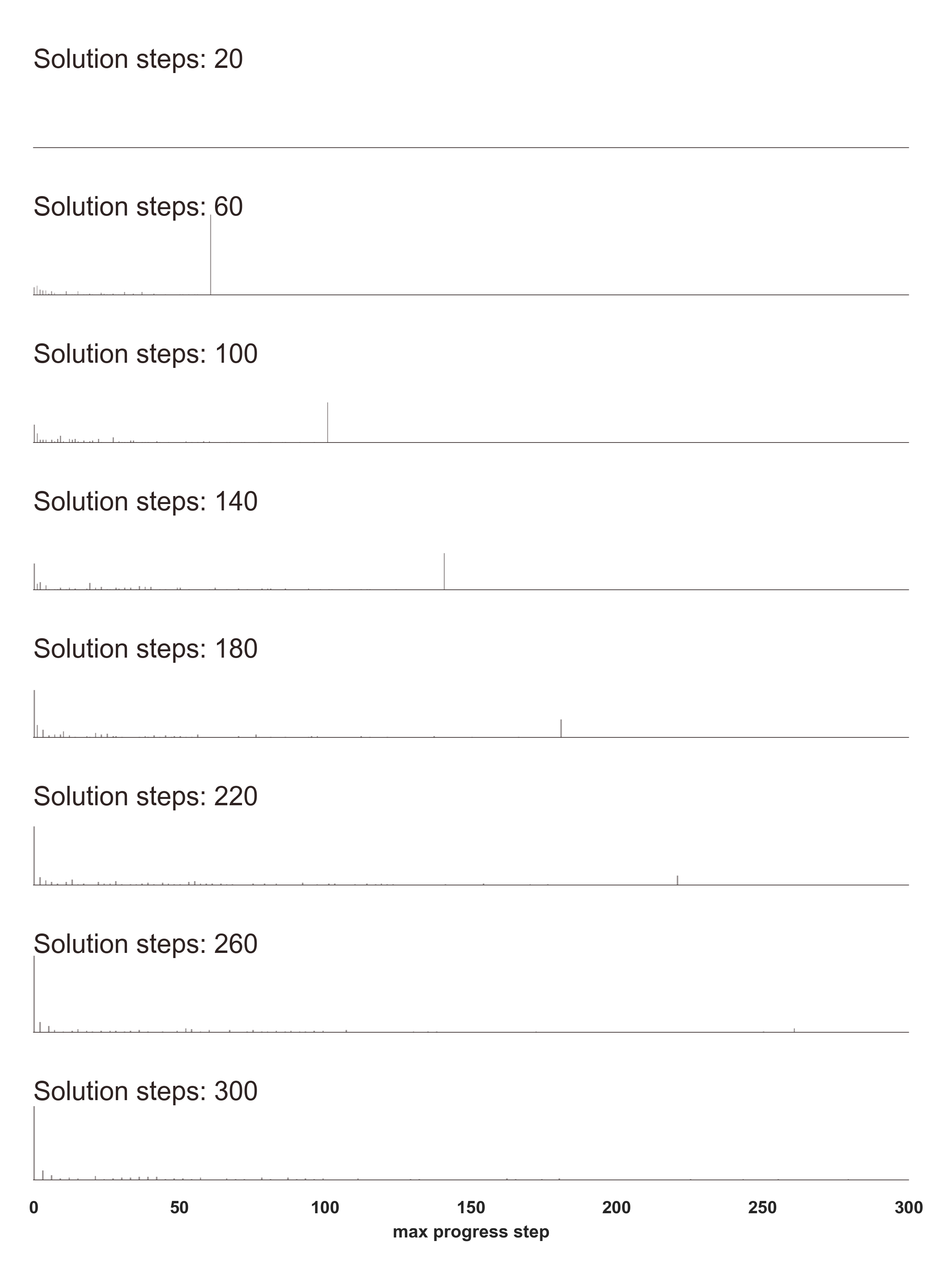}
  \caption{Distribution of first-violation steps for Gemini-2.5-Flash across varying logical depths ($L$). As $L$ (total required path length) increases, the distribution of first errors tends to shift leftward, indicating that models are more likely to fail at earlier steps in longer problems. This suggests that anticipated global complexity impacts reasoning from the outset. Experimental parameters in this figure are the same as those in Figure 1.}
  \label{fig:S_fail_dist_vs_L}
\end{figure*}

Figure~\ref{fig:S_fail_dist_vs_L} shows how the distribution of these first-error positions shifts with the overall problem complexity, represented by logical depth ($L$). As detailed in the main paper, an increase in $L$ tends to cause errors to occur earlier in the reasoning chain.

\begin{figure*}[hbt!]
  \centering
  \includegraphics[width=1.0\linewidth]{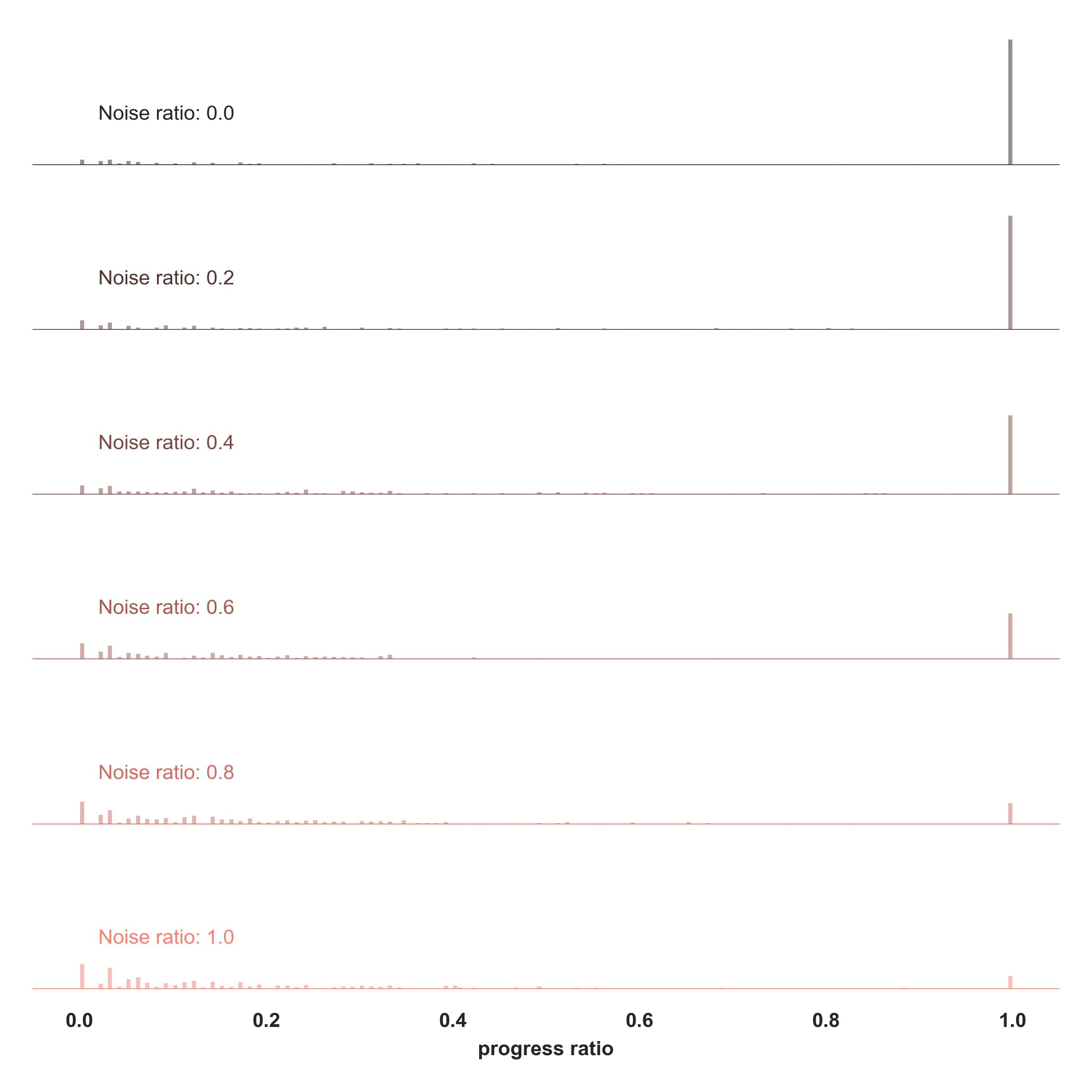}
  \caption{Impact of increasing noise ratio on the distribution of failure steps for Gemini 2.5 Flash. As noise (proportion of distracting facts) increases, failures tend to occur earlier in the reasoning chain. This reflects increased difficulty in isolating relevant information and maintaining focus. Fixed experimental parameters in this figure are the same as those in Figure 1.}
  \label{fig:S_dist_vs_noise}
\end{figure*}

Similarly, Figure~\ref{fig:S_dist_vs_noise} illustrates how the introduction of contextual noise (distracting facts) affects the point of failure. Increased noise also tends to precipitate earlier errors in the reasoning sequence, as discussed in the main paper in relation to sensitivity to noise (Section~\ref{sec:complexity_dimensions}) and its impact on error patterns (Section~\ref{sec:error_pattern}).

\section{Supplementary Figures} 
\label{app:dataset_figures}
This appendix provides supplementary figures that offer further visual support for analyses presented in the main paper. These figures illustrate the impact of various complexity dimensions and provide comparative views of model performance, elaborating on points made throughout Section~\ref{sec:results} (Benchmarking Results) of the main paper.

\begin{figure*}[hbt!]
  \centering
  \includegraphics[width=1.0\linewidth]{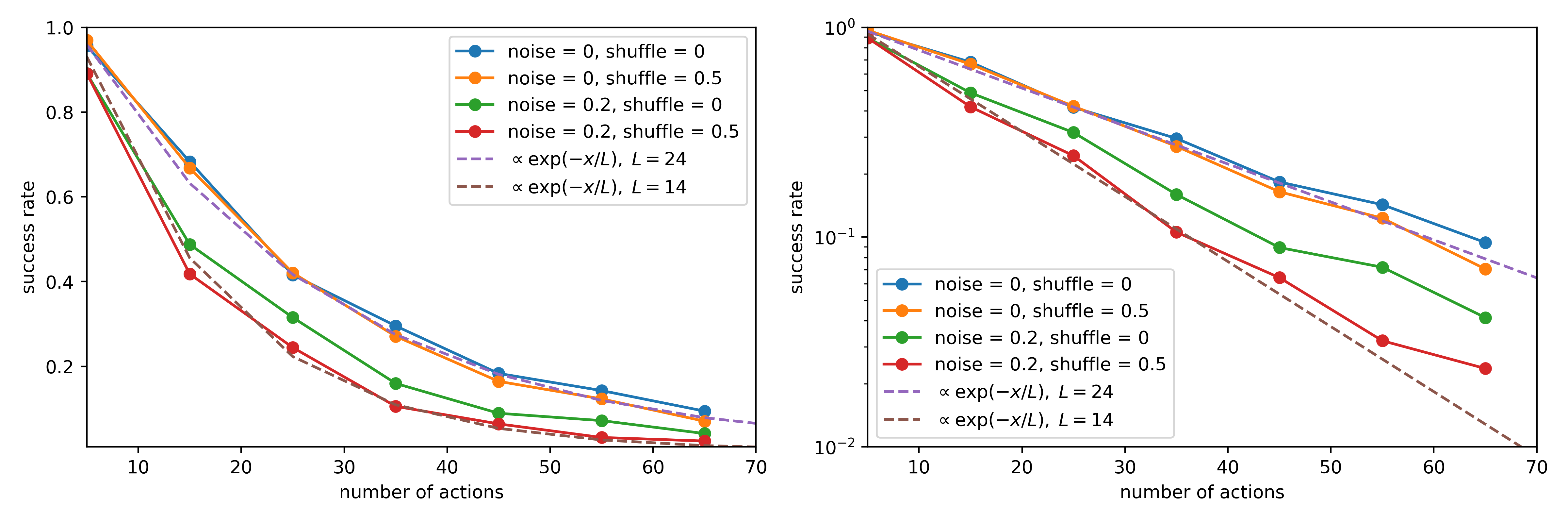}
  \caption{Pass@1 success rate for Llama-4 Maverick-17B-128E-Instruct versus solution length ($L$) under different noise and shuffle ratios. Left: Linear scale. Right: Log-linear scale. Performance degrades with increased noise but is less affected by shuffle ratios. Fixed experimental parameters in this figure are the same as those in Figure 1.}
  \label{fig:S_noise_shuffle_perf_vs_steps}
\end{figure*}

Figure~\ref{fig:S_noise_shuffle_perf_vs_steps} details the performance of Llama-4 Maverick-17B-128E-Instruct under varying levels of noise and fact shuffling. This supports the discussion in the main paper (Section~\ref{sec:complexity_dimensions}, on how these factors, especially in combination, affect success rates, with noise being a dominant factor.

\begin{figure*}[hbt!]
  \centering
  \includegraphics[width=1.0\linewidth]{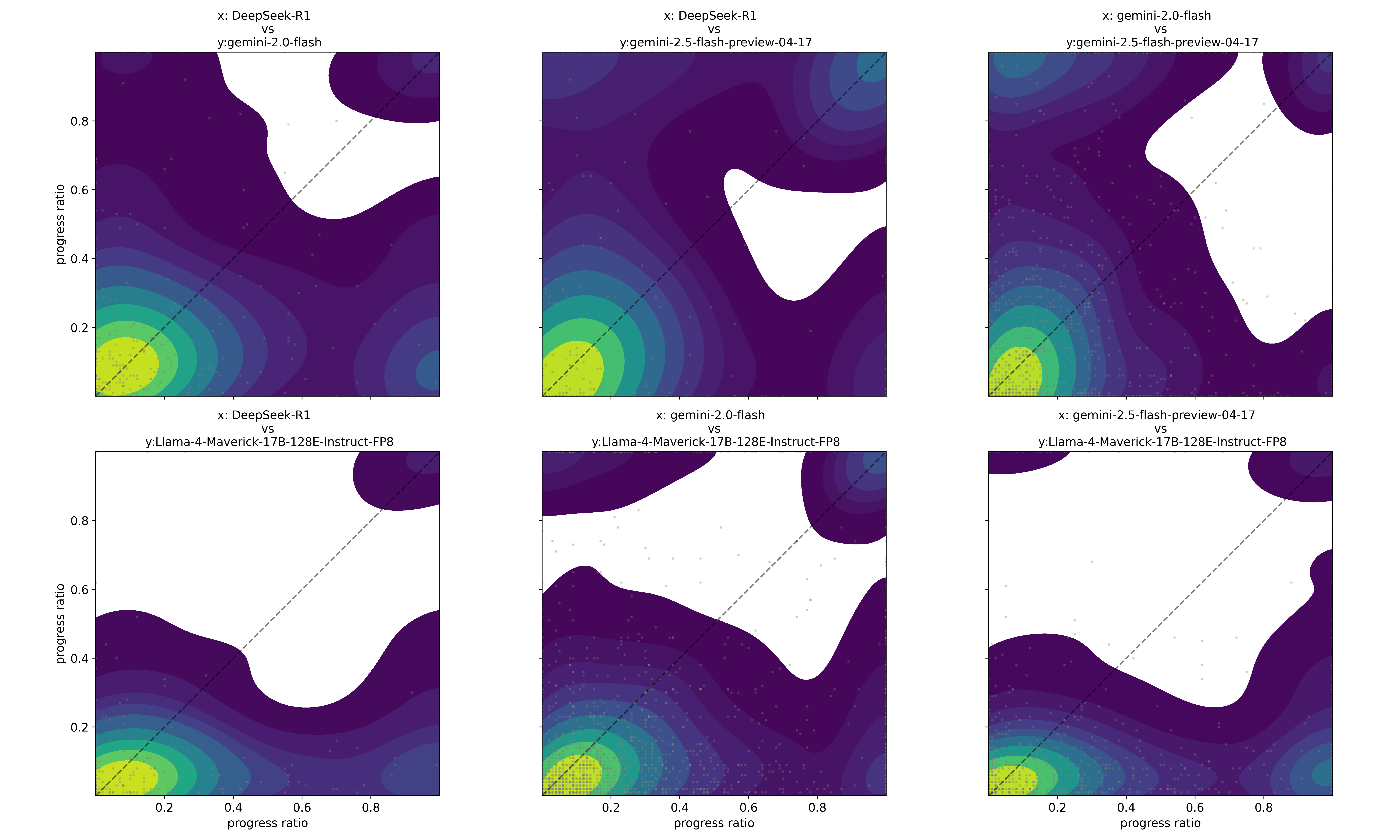}
  \caption{Scatter and density plots of progress ratios per task instance, comparing model pairs on the tasks. These plots illustrate performance agreement and disparities on the same instances of pathfinding tasks. Notably, Gemini-2.5-Flash (example) often succeeds on instances where other models achieve near-zero progress. Data from experiments in Figure~\ref{fig:universal_vs_L} (main paper).}
  \label{fig:S_progress_vs_progress}
\end{figure*}

To illustrate the performance consistency and disparities across different models, as detailed in Section~\ref{sec:hierarchy}, Figure~\ref{fig:S_progress_vs_progress} presents scatter and density plots of mean progress ratios. These plots clearly demonstrate that model performance hierarchies are not strictly linear. They reveal 'performance inversions'—instances, also noted in Section~\ref{sec:hierarchy}, where models with typically lower overall performance (e.g., lower average $L_0$) occasionally solve specific complex problems that models with higher average $L_0$ values fail on.

\begin{figure*}[hbt!]
  \centering
  \includegraphics[width=1.0\linewidth]{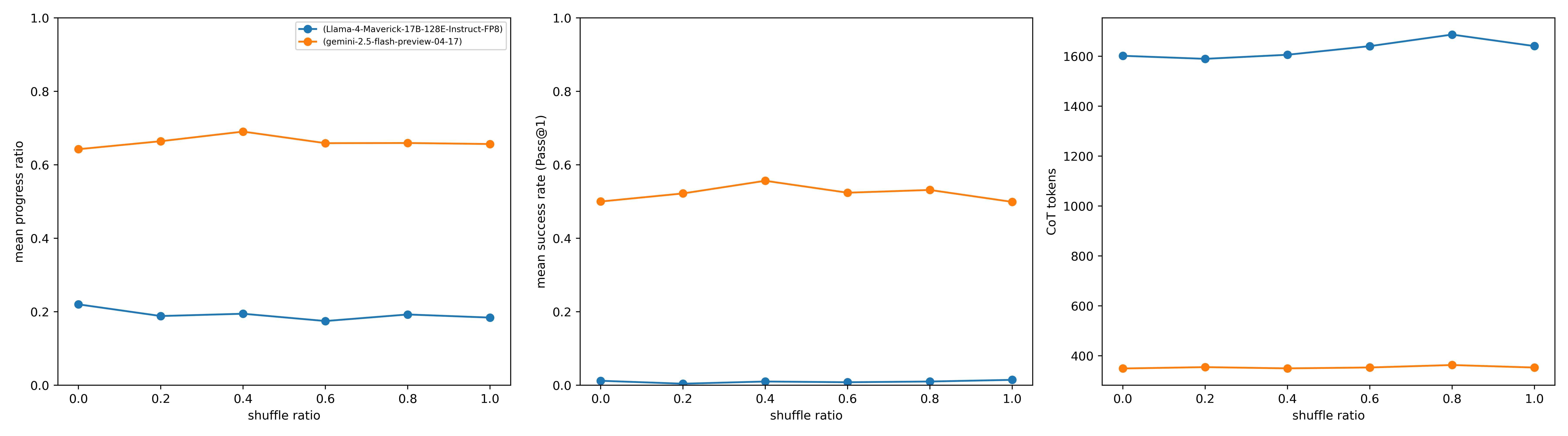}
    \caption{Impact of shuffle ratio on Pass@1 success rate. Varying the degree of mixing (shuffle) between supporting and distracting facts shows minimal impact on performance for Gemini 2.5 Flash and Llama-4 Maverick, suggesting robustness to fact order when noise is controlled. The generation and sampling of maze instances for these tasks follow the same methodology detailed for experiments in the main paper (Figures \ref{fig:vs_backtracking} and~\ref{fig:vs_noise}).}
  \label{fig:S_vs_shuffle}
\end{figure*}

Figure~\ref{fig:S_vs_shuffle} isolates the impact of shuffle ratio on model performance when other factors like noise are controlled. This visualization corresponds to the findings discussed in the main paper (Section~\ref{sec:complexity_dimensions}, "Fact Ordering (Shuffle Ratio)") that simple reordering of facts has a minimal impact on the performance of the evaluated models under low-noise conditions.

\begin{figure*}[hbt!]
  \centering
  \includegraphics[width=1.0\linewidth]{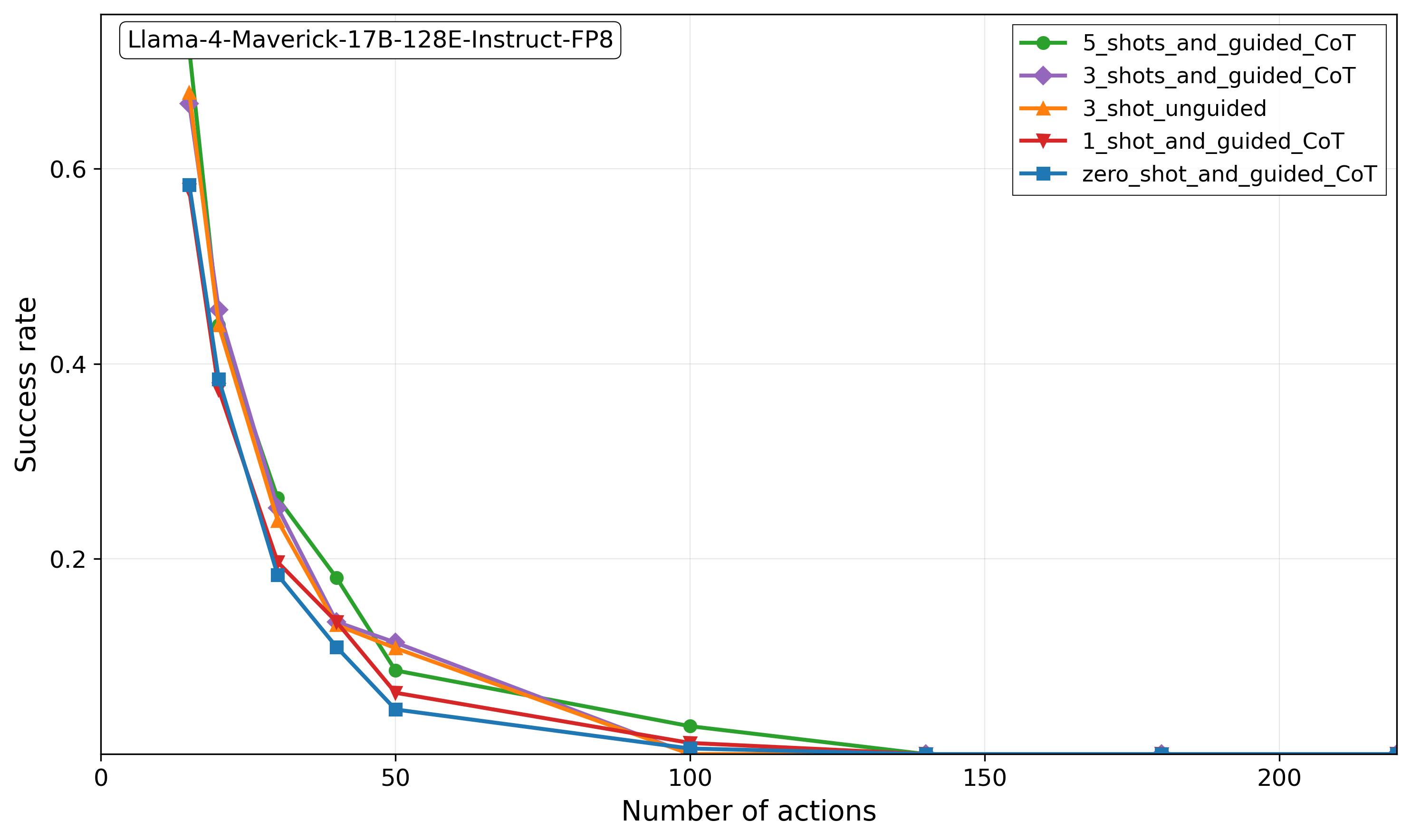}
    \caption{The impact of including different number of reference examples in the prompt as part of in-context learning. Increasing the number of examples leads to slight improvements in performance. The experimental parameters used here are the same as ones in Figure 1.}
  \label{fig:few_shot_abalation}
\end{figure*}

Figure~\ref{fig:few_shot_abalation} isolates the impact of adding more examples in the instruction prompt, showing a clear improvement once more than a single example is included compared to using none or only one.

Figure~\ref{fig:gpt5} is added in this revised version of the supplementary section to reflect that even the most recent SOTA models released by OpenAI suffer from the same performance drop observed in the main paper.

\begin{figure*}[hbt!]
  \centering
  \includegraphics[width=1.0\linewidth]{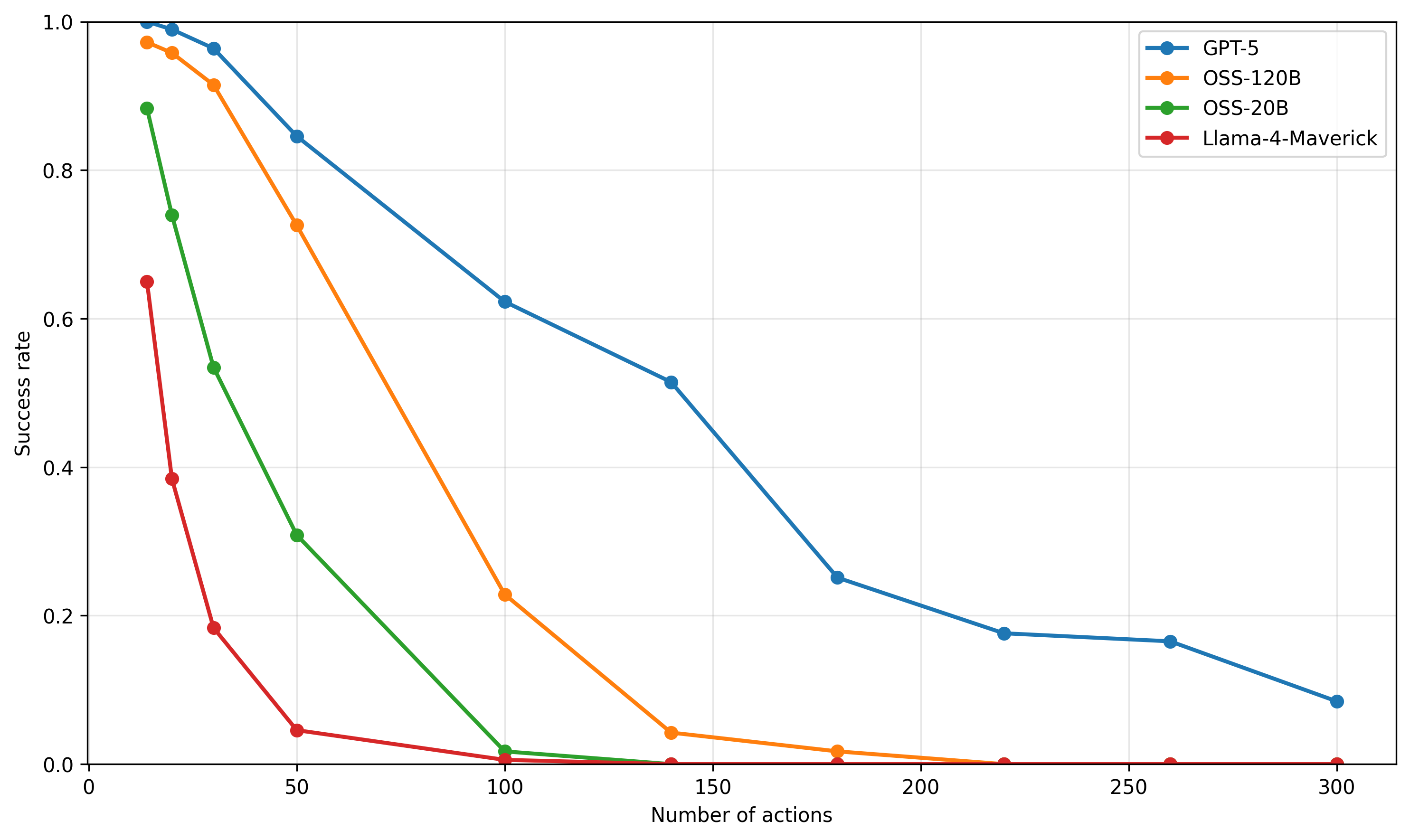}
    \caption{This figure is added to reflect that the recent closed (GPT-5) and open sourced models (OSS-20B/120B) released by OpenAI also follow the same universal failure patterns highlighted in this paper. The data used here as well as experimental settings is the same as the one used in Figure 1 of the main paper. We include Llama-4-Maverick which is also used in Figure 1 as the benchmark reference.}
  \label{fig:gpt5}
\end{figure*}